\documentclass[journal, twoside]{IEEEtran}
                           
\usepackage{graphicx}
\usepackage{picinpar}
\usepackage{amsmath,bm}
\usepackage{url}
\usepackage{colortbl}
\usepackage{soul}
\usepackage{multirow}
\usepackage{pifont}
\usepackage{color}
\usepackage{alltt}
\usepackage[hidelinks]{hyperref}
\usepackage{enumerate}
\usepackage{siunitx}
\usepackage{breakurl}
\usepackage{epstopdf}
\usepackage{pbox}
\usepackage{indentfirst}
\usepackage{mathtools}
\usepackage{booktabs}
\usepackage{makecell}
\usepackage{subfigure}
\usepackage{float}
\usepackage{threeparttable}
\usepackage{algorithmic}
\usepackage{algorithm}
\usepackage{amsfonts}
\usepackage[allow-number-unit-breaks]{siunitx}
\usepackage[dvipsnames]{xcolor}
\usepackage{cite}

\newcommand{\etalcite}[1]{et al.~\cite{#1}}

\begin{document}

\title{Wheel-GINS: A GNSS/INS Integrated Navigation System with a Wheel-mounted IMU}

\author{Yibin~Wu$^{1}$~Jian~Kuang$^{2}$~Xiaoji~Niu$^{2}$~Cyrill~Stachniss$^{1}$~Lasse~Klingbeil$^{1}$~Heiner~Kuhlmann$^{1}$
\thanks{$^{1}$Yibin Wu, Cyrill Stachniss, Lasse Klingbeil, and Heiner Kuhlmann are with the Center for Robotics and Institute of Geodesy and Geoinformation, University of Bonn, Bonn, Germany. Cyrill Stachniss is additionally with the Department of Engineering Science at the University of Oxford, UK, and with the Lamarr Institute for Machine Learning and Artificial Intelligence, Germany, {\{{\tt\small firstname.lastname}\}{\tt\small @igg.uni-bonn.de}}.}
\thanks{$^{2}$Jian Kuang, and Xiaoji Niu are with the GNSS Research Center, Wuhan University, Wuhan, China, {\{{\tt\small kuang, xjniu}\}{\tt\small @whu.edu.cn}}. (Corresponding Author: \textit{Jian Kuang})}
\thanks{This work has partially been funded by the Fundamental Research Funds for the Central Universities (2042023kf0124) and Hubei Provincial Natural Science Foundation Program (2023AFB021).}
}

\markboth{IEEE Transactions on Intelligent Transportation Systems. Preprint Version. January, 2025}{Wu \MakeLowercase{\textit{et al.}}: Wheel-GINS} 

\maketitle

\begin{abstract}
A long-term accurate and robust localization system is essential for mobile robots to operate efficiently outdoors. Recent studies have shown the significant advantages of the wheel-mounted inertial measurement unit (Wheel-IMU)-based dead reckoning system. However, it still drifts over extended periods because of the absence of external correction signals. 
To achieve the goal of long-term accurate localization, we propose Wheel-GINS, a Global Navigation Satellite System (GNSS)/inertial navigation system (INS) integrated navigation system using a Wheel-IMU. Wheel-GINS fuses the GNSS position measurement with the Wheel-IMU via an extended Kalman filter to limit the long-term error drift and provide continuous state estimation when the GNSS signal is blocked. Considering the specificities of the GNSS/Wheel-IMU integration, we conduct detailed modeling and online estimation of the Wheel-IMU installation parameters, including the Wheel-IMU leverarm and mounting angle and the wheel radius error. Experimental results have shown that Wheel-GINS outperforms the traditional GNSS/Odometer/INS integrated navigation system during GNSS outages. At the same time, Wheel-GINS can effectively estimate the Wheel-IMU installation parameters online and, consequently, improve the localization accuracy and practicality of the system. The source code of our implementation is publicly available (https://github.com/i2Nav-WHU/Wheel-GINS).
\end{abstract}


\section{Introduction}
\begin{figure}[t]
	\centering
	\setlength{\abovecaptionskip}{0.cm}
	\includegraphics[width=8.8cm]{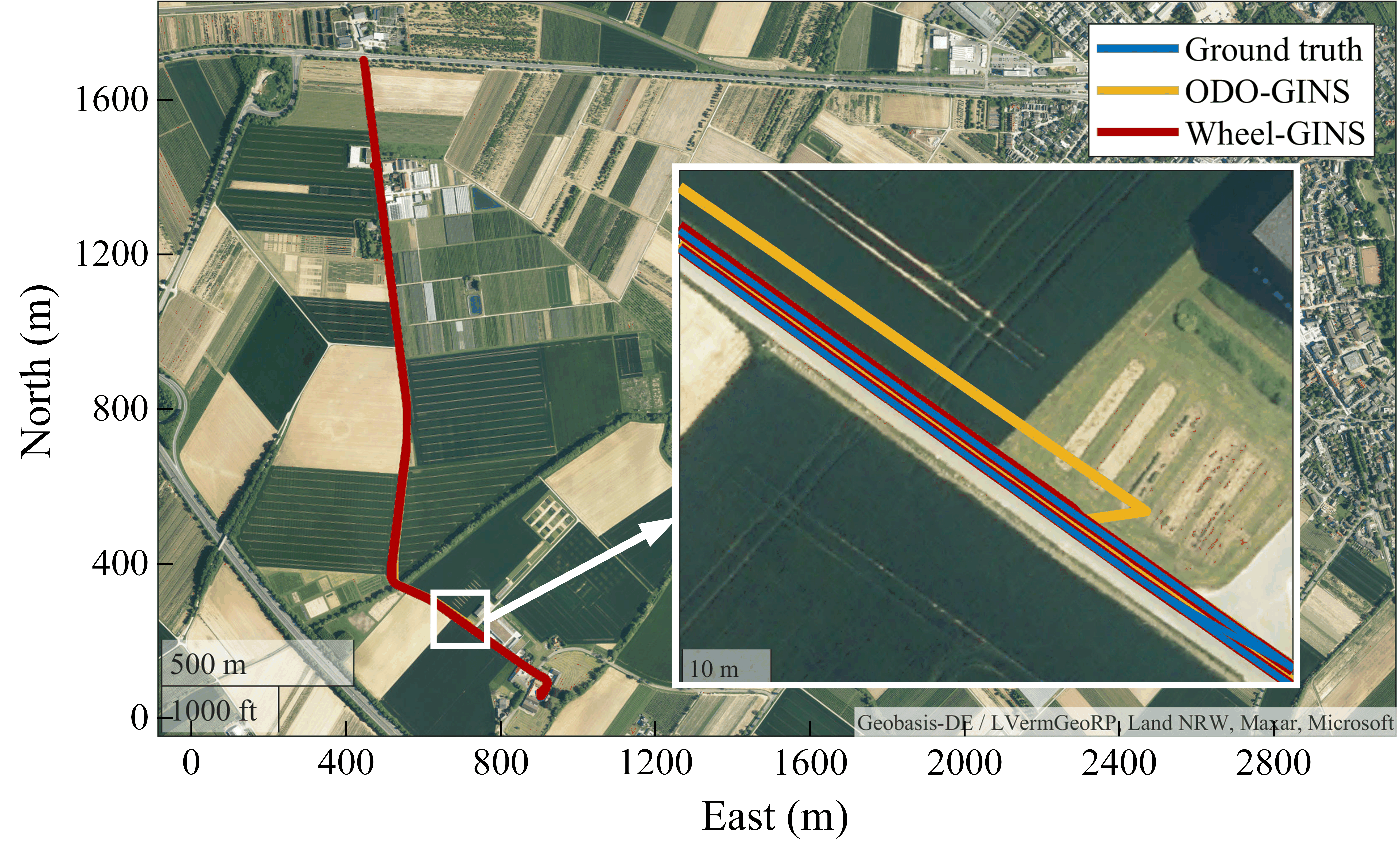}
	\caption{The vehicle trajectories estimated by the proposed Wheel-GINS and the conventional GNSS/Odometer/INS integrated navigation system (ODO-GINS) against the ground truth in our car experiment. The car traversed the same road back and forth twice. The enlarged view shows the position estimates during \SI{120}{s} GNSS outage. We can see that Wheel-GINS has significantly improved the positioning accuracy compared to ODO-GINS during GNSS outages.}
	\label{fig:cover_fig}
\end{figure}

\IEEEPARstart{L}{ocalization} is the foundation for many other tasks in mobile robots, such as environment perception~\cite{wong2023tits,xiaozhu2024tmc,wong2024cvpr,pan2024tro} and path planning~\cite{lavalle2006planning}. An accurate and robust localization system operating over a long period is crucial for the robots to function effectively without human supervision and intervention. To this end, modern robots are always equipped with multiple sensors, e.g., LiDAR, camera, odometer, inertial measurement unit (IMU), Global Navigation Satellite System (GNSS) receiver, etc., to take their complementary advantages via sensor fusion for state estimation. Among all these sensors, inertial sensors play an essential role because they are self-contained. In other words, they measure the robot's egomotion independently without other external signals and interaction with the environment. In addition, inertial sensors are cheap and consume lightweight computational resources. Therefore, investigating the inertial navigation system (INS) offers significant potential for enhancing the performance of sensor fusion-based navigation systems.

In two recent papers~\cite{niu2021, wu2021}, the authors proposed Wheel-INS, a 2D dead reckoning system using only one wheel-mounted IMU (Wheel-IMU). To be specific, they multiply the gyroscope readings of the Wheel-IMU with the wheel radius to compute the wheel velocity, which is then combined with the non-holonomic constraints to integrate with the strapdown INS~\cite{savage2007strapdown, Shin2005} via an extended Kalman filter (EKF). Wheel-INS offers two key advantages. First, it uses the Wheel-IMU gyroscope measurement to calculate wheel speed, achieving similar information fusion as the traditional odometer-aided INS with only one sensor. Second, the continuous rotation of the wheel helps mitigate the constant IMU bias errors. Experimental results~\cite{niu2021} show that Wheel-INS outperforms the traditional odometer-aided INS in terms of pose accuracy. In addition, it exhibits significant resilience to constant IMU bias error compared to the traditional odometer-aided INS.

While Wheel-INS demonstrates effective dead reckoning performance, it is a relative positioning system that cannot constrain long-term error drift. To address this limitation, Wu~et~al. proposed Wheel-SLAM~\cite{wu2022ral}, which used the road bank angle, indicated by the vehicle roll angle estimation from Wheel-INS, to construct a terrain map. This enables loop closure detection and pose correction to limit the error accumulation. However, Wheel-SLAM is only applicable for those applications where the robot moves in constrained environments while failing in scenarios where the robot does not have the opportunity to revisit the places it has been before, for example, a self-driving car driving from one city to another. Therefore, external correction signals are necessary to limit the long-term error drift of Wheel-INS.

\begin{figure}[t]
	\centering
	\includegraphics[width=8.8cm]{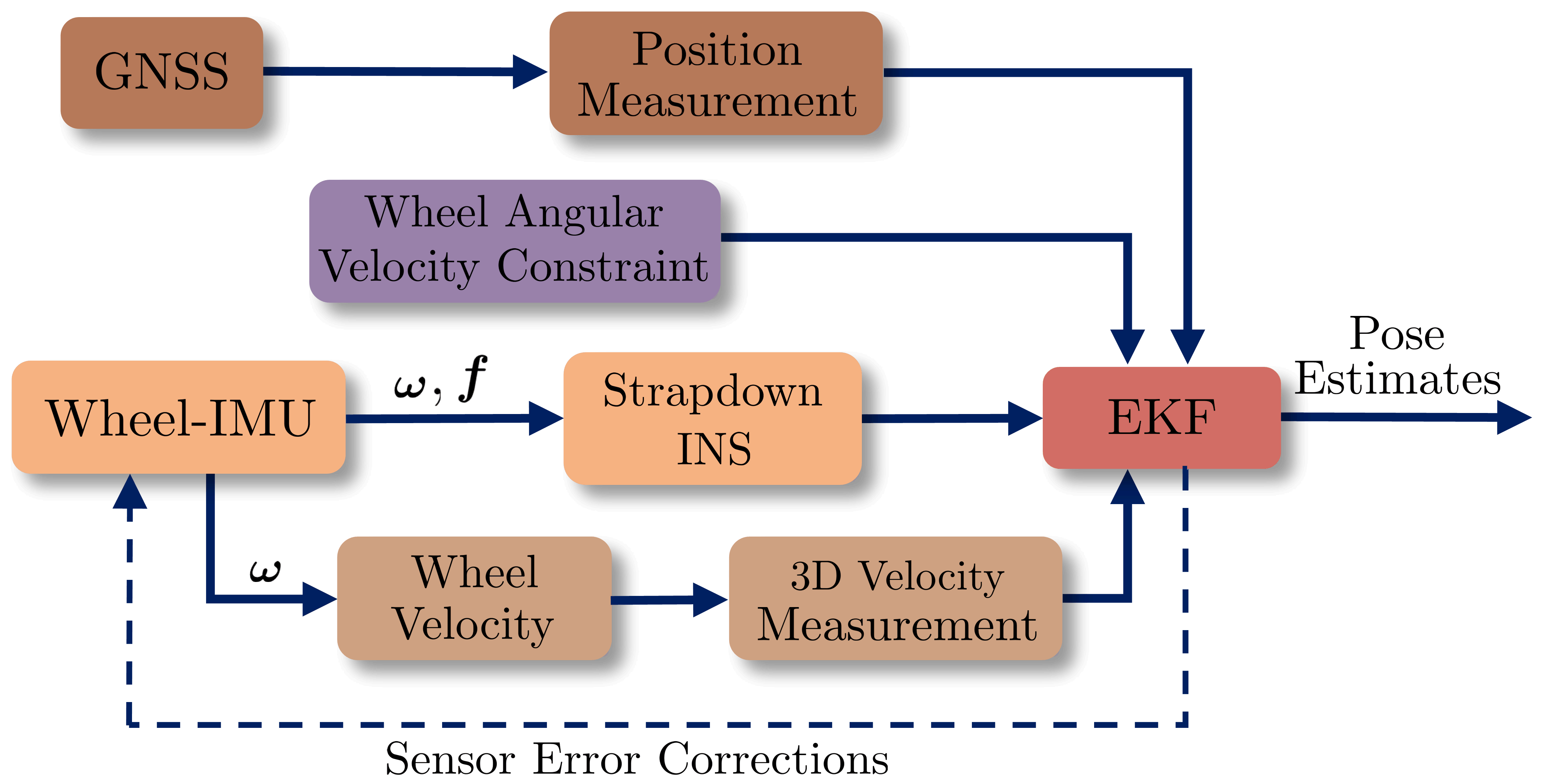}
	\caption{System overview of Wheel-GINS. $\bm\omega$ and $\bm{f}$ are the angular velocity and specific force measured by the Wheel-IMU, respectively.}
	\label{fig:wheelgins_overview}
\end{figure}

Because of the complementary advantages of GNSS and INS, the GNSS/INS integrated navigation system has been widely applied to real-world robot navigation applications. On the one hand, GNSS~\cite{li2022sana} can provide accurate absolute positions to correct the error drift of INS. On the other hand, INS can provide continuous state estimation when GNSS is not available. Then, a natural question arises: \textit{How to integrate GNSS with Wheel-INS to achieve the goal of a long-term accurate and robust navigation system?}  

The main contribution of this paper is the development of an algorithm to fuse GNSS with Wheel-INS, along with a study of its characteristics and an evaluation of its performance. To the best of our knowledge, this is the first GNSS/INS integrated navigation system in the literature that uses a wheel-mounted IMU. Specifically, we integrate the GNSS position measurement into the EKF pipeline building on top of Wheel-INS; thus, Wheel-GINS can achieve a similar information fusion scheme as the conventional GNSS/Odometer/INS integrated navigation system (ODO-GINS). However, Wheel-GINS is more promising than ODO-GINS due to the advantages of Wheel-INS. Fig.~\ref{fig:cover_fig} shows the vehicle trajectories estimated by the proposed Wheel-GINS and ODO-GINS compared to ground truth in our car experiment. We can see that the error drift of Wheel-GINS is significantly smaller compared to ODO-GINS during GNSS outages.

In addition, the misalignment between the GNSS and the Wheel-IMU is different from that between the GNSS and the normal IMU, which is usually placed on the top or in the body of the robot. To handle this issue and improve the practicality of the proposed Wheel-GINS, we perform detailed modeling of the Wheel-IMU installation parameters, including the Wheel-IMU leverarm (position misalignment) and mounting angle (attitude misalignment), and the wheel radius scale error to estimate them online. Furthermore, we propose a wheel angular velocity constraint model to accelerate the convergence of the Wheel-IMU mounting angle online estimation. Details of the Wheel-IMU installation parameters are presented in Section III-A. Fig.~\ref{fig:wheelgins_overview} overviews the algorithm structure of the proposed Wheel-GINS.

In sum, we make two key claims: (i) Wheel-GINS has significantly reduced the position error drift compared to ODO-GINS during GNSS outages; (ii) Wheel-GINS can effectively estimate the Wheel-IMU installation parameters, including the Wheel-IMU leverarm and mounting angle and the wheel radius scale error online, thus improving the pose estimation accuracy. Our experimental evaluation backs up these claims. 


\section{Related Work}
\subsection{State Estimation using Wheel-mounted IMUs}
Researchers have proposed various low-cost alternatives to wheel odometers for measuring wheel velocity using wheel-mounted IMUs~\cite{naser2021sensors, gersdorf2013, coulter2011}. Youssef~\etalcite{naser2021sensors} proposed detecting peaks and valleys in the accelerometer signal to count complete wheel cycles and compute the traveled distance of the robot. Coulter~\etalcite{coulter2011} used the Wheel-IMU accelerometer to measure the wheel's rotation angle and movement duration. Gersdorf~\etalcite{gersdorf2013} employed a gyroscope on the wheel's rotation axis and two accelerometers on the wheel plane to measure vehicle acceleration and estimate wheel velocity via EKF. However, these studies focus solely on vehicle velocity information, not vehicle pose.

Collin~\etalcite{collin2014tim, collin2014tvt} proposed the first 2D localization system based only on a Wheel-IMU. It used the two accelerometers on the wheel plane to estimate the wheel rotation angle and, thus, the traveled distance. Meanwhile, it used the gyroscope measurements to calculate the vehicle heading. However, this method assumed a constant vehicle speed, and the misalignment error was not considered. Recently, Tan~\cite{tan2024sensorj} proposed to use the tri-axis accelerometer output and the gyroscope output at the rotation axis of the Wheel-IMU as the observation model to fuse with the kinematic model of the wheel angular velocity and acceleration via EKF.

Niu~\etalcite{niu2021} proposed an EKF-based approach to estimate the 2D robot state using only a Wheel-IMU. They obtained the wheel velocity by multiplying the gyroscope readings on the wheel rotation axis with the wheel radius instead of relying on an odometer or wheel encoder. This velocity was then fused with the non-holonomic constraint to integrate with the strapdown INS via an EKF, achieving a methodology akin to traditional odometer-aided INS approaches~\cite{wu2009}. Based on Wheel-INS, the authors further compared three different observation models obtained from the Wheel-IMU~\cite{wu2021} and extended single Wheel-IMU to multiple IMUs mounted on different places of the vehicle for better state estimation performance~\cite{wu2022tits}. To limit the error drift of Wheel-INS, they proposed Wheel-SLAM~\cite{wu2022ral}, using the vehicle roll angle estimated by Wheel-INS to build a terrain map for loop closure detection and pose correction.

However, current studies have only investigated the dead reckoning system using Wheel-IMU. The feasibility, system characteristics, and performance of fusing the Wheel-IMU with global positioning sensors, such as GNSS, have not been studied so far.

\subsection{GNSS/INS Fusion}
Although GNSS/Wheel-IMU fusion has not been investigated, the GNSS/INS integrated navigation system, using a normal IMU mounted on the body of the vehicle, has been extensively studied in the past decades~\cite{angrisano2010phd, wen2021navigation, zhangquan2024tiv, meng2024tim, xiaozhu2020tits}. Nowadays, it has become a standard component of many vehicle navigation systems. GNSS/INS integration framework can be divided into loosely coupled and tightly coupled, depending on which measurement from GNSS is used. In the loosely coupled system~\cite{Shin2005, zhang2022ral, hua2023sana}, the position (and velocity) estimated by the GNSS receiver is fused with INS, while the tightly coupled system~\cite{wen2019tvt, wang2020tim, xiaozhu2022tits} directly integrates raw GNSS measurements, such as pseudorange and carrier phase measurements. Although various methods, such as factor graphs~\cite{loeliger2004factorgraph, suzuki2024ral, zhang2024tiv}, have been proposed to fuse GNSS with INS, the most common one is still the EKF~\cite{wen2021navigation, Shin2005}. Because this study focuses on investigating the idea and feasibility of fusing GNSS with Wheel-IMU instead of algorithm research for GNSS/INS fusion, we adopt the classical loosely coupled framework and use the EKF in the proposed Wheel-GINS.

To improve the accuracy of GNSS/INS integrated navigation system during GNSS outages, various sensors, such as cameras~\cite{niu2023ral, chi2023ral,li2022ral}, LiDARs~\cite{liu2024tiv, li2022iot, chang2024tiv}, and odometers~\cite{Shin2005, niu2007nav, Zhang2021mst} have been introduced into the GNSS/INS integrated system. Unlike cameras and LiDARs, which depend on environmental conditions, odometers are preferable due to their independence from the environment~\cite{wu2009, ouyang2020}. The vehicle velocity provided by the odometer is always combined with the non-holonomic constraints~\cite{dissanayake2001} as a 3D velocity measurement to fuse with INS. It has been illustrated that the odometer and non-holonomic constraints significantly enhance the pose accuracy when GNSS is not available~\cite{zhangquan2020, martin2020tiv}. Because the proposed Wheel-GINS achieves a similar information fusion as the traditional GNSS/Odometer/INS integrated navigation system, we use it as a benchmark to illustrate the performance of Wheel-GINS in the experiments. 

\section{Prerequisites}

\begin{figure}[t]
	\centering
	\includegraphics[width=8.8cm]{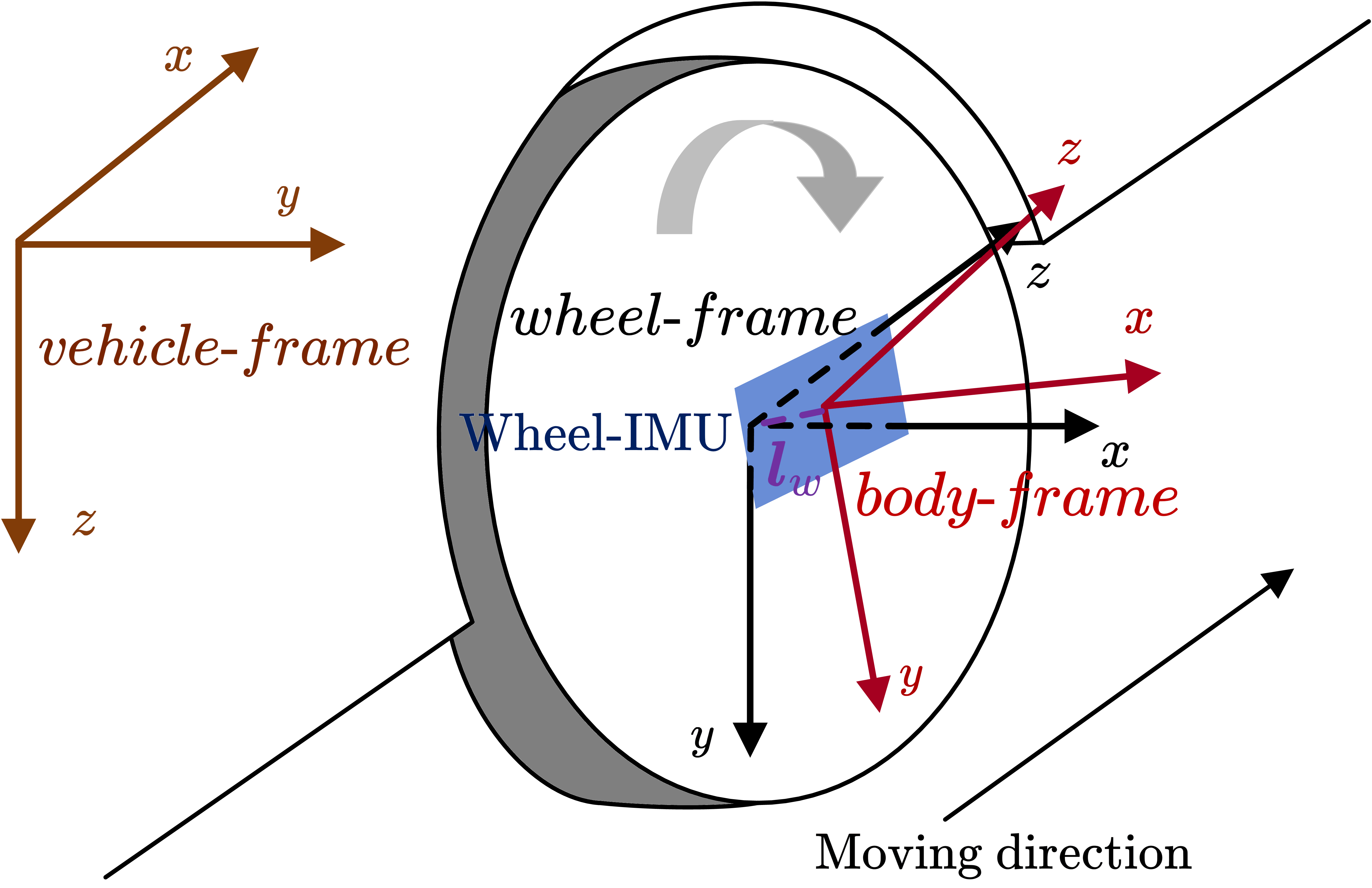}
	\caption{Illustration of the \textit{vehicle}-frame, \textit{wheel}-frame, and IMU \textit{body}-frame~\cite{niu2021}. More details can be found in Table~\ref{tab:coordinates}. $\boldsymbol{l}_w$ indicates the leverarm between the Wheel-IMU and the wheel center; the non-parallelism of the \textit{wheel}-frame and \textit{body}-frame indicates the mounting angle between the Wheel-IMU and the wheel. Note that the origin of the \textit{vehicle}-frame is defined at the wheel center (see Table~\ref{tab:coordinates}). Here, we plot it on the vehicle body only for easy visualization.}
	\label{fig:coordinates}
\end{figure}

\subsection{Sensors Setup}
In Wheel-GINS, only two sensors, a GNSS module and a wheel-mounted IMU are used. The IMU is mounted on the vehicle's non-steering wheel, while the GNSS antenna is placed on the top of the vehicle. Our Wheel-IMU design uses an in-built battery for power supply and a Bluetooth module for data transmission. Table~\ref{tab:coordinates} lists the definitions of the relevant coordinate systems. Fig.~\ref{fig:coordinates} provides a detailed depiction of the vehicle and IMU coordinate systems. 

Given a stable vehicle structure, the difference between the wheel heading and vehicle heading should remain constant, according to the definitions of the \textit{vehicle}-frame and \textit{wheel}-frame, namely,
\begin{equation}
\psi_{w} = \psi_{v} + \pi/2
\label{headingdifference},
\end{equation}
where $\psi_{w}$ and $\psi_{v}$ denote the wheel heading and vehicle heading, respectively. If the Wheel-IMU were perfectly aligned with the wheel, its heading would equal the wheel heading, and consequently, we could directly determine the vehicle's heading from the Wheel-IMU state. However, misalignment is inevitable in practical systems that integrate information from multiple sensors. As shown in Fig.~\ref{fig:coordinates}, the misalignment errors of Wheel-IMU include position misalignment and attitude misalignment. In Wheel-INS~\cite{niu2021}, the authors illustrated the negative impact of misalignment errors on the system performance. However, in Wheel-INS, the Wheel-IMU misalignment errors are calibrated offline, which causes inconvenience for practical use. In Wheel-GINS, we incorporate the attitude and position misalignment errors between the Wheel-IMU and the wheel, along with errors in wheel radius, into the state vector. These parameters are estimated online to enhance the system's practicality.

The position misalignment is also termed as leverarm ($\bm{l}_{w}$), indicating the vector pointing from the Wheel-IMU center to the wheel center expressed in the \textit{body}-frame. We take only the components in the \textit{y}-axis and \textit{z}-axis into consideration. The reason is threefold. First, the leverarm error on the wheel plane is more important. Second, the leverarm error in the direction of the rotation axis is trivial if the attitude misalignment is compensated. Third, reducing one dimension in the state vector can save some computation resources. Therefore, we have
\begin{equation}
\bm{l}_{w} = 
\begin{bmatrix}
    l_y & l_z
\end{bmatrix}^{\top}.
\end{equation}

The attitude misalignment is also termed as mounting angle ($\bm{\phi}_{m}$), which can be represented by a set of Euler angles $\varphi_m$ (roll mounting angle), $\theta_m$ (pitch mounting angle), and $\psi_m$ (heading mounting angle). With this, the \textit{wheel}-frame can be rotated by $\psi_m$, $\theta_m$, and $\varphi_m$ around the \textit{z}-axis, \textit{y}-axis, and \textit{x}-axis respectively to align with the \textit{body}-frame. Because the wheel continuously rotates with the \textit{x}-axis, the roll mounting angle is negligible according to the definition of the \textit{wheel}-frame. Therefore, we only account for the pitch mounting angle and the heading mounting angle, namely,
\begin{equation}
\bm{\phi}_{m} = 
\begin{bmatrix}
    \theta_m & \psi_m
\end{bmatrix}^{\top}.
\end{equation}

Note that we consider the 3D vector of both the leverarm and mounting angle when we derive the observation model equations to simplify the expression. In practice, we extract the components from the equations corresponding to the actual misalignment error to construct the matrix for the final computation in the EKF (see Section IV-B).

In addition, the measured wheel radius may also have an error because the vehicle weight, temperature, and tire pressure would all cause wheel deformation. It is necessary to compensate for this error because it plays an important role in computing the wheel velocity in Wheel-GINS. Here, we model the wheel radius error as a scale factor error, namely,

\begin{equation}
\hat{r} = (1+s_r)r,
\end{equation}
where $\hat{r}$ is the measured wheel radius; $s_r$ is the wheel radius scale error, augmented into the state vector for online estimation; $r$ is the true wheel radius, which is unknown. More details of the online estimation of the Wheel-IMU installation parameters are presented in Section IV.

\begin{table}[t]
	\renewcommand\arraystretch{1.5}
	\centering
	\caption{Definitions of the Coordinates Systems in Wheel-GINS}
	\label{tab:coordinates}
	\begin{tabular}{p{1.6cm}p{2.8cm}p{3.2cm}}
		\toprule
		\multicolumn{1}{c}{Symbol} & \multicolumn{1}{c}{Description} & \multicolumn{1}{c}{Definition} \\
		\midrule
            {\textit{earth}-frame} & {The Earth-Centered Earth-Fixed (ECEF) coordinates system.} & {\textit{origin}: the center of mass of the Earth.\newline
			\textit{x-axis}: toward the mean meridian of Greenwich.\newline
			\textit{y-axis}: completing a right-handed orthogonal frame.\newline
			\textit{z-axis}: parallel to the mean spin axis of the Earth.}
		\\
            {\textit{n}-frame} & {The navigation frame.} & {\textit{origin}: the same as \textit{body}-frame.\newline
			\textit{x-axis}: north.\newline
			\textit{y-axis}: east.\newline
			\textit{z-axis}: downward vertically.}
		\\
		{\textit{vehicle}-frame} & {The coordinates system of the vehicle.} & 
			{\textit{origin}: center of the wheel where the IMU is placed.\newline
			\textit{x-axis}: forward.\newline
			\textit{y-axi}s: right.\newline
			 \textit{z-axis}: down.}
		\\
            {\textit{wheel}-frame} & {The coordinates system of the wheel.} & 
			{\textit{origin}: wheel center.\newline
			\textit{x-axis}: right, perpendicular to the wheel plane.\newline
			\textit{y}- and \textit{z-axes} are parallel to the wheel plane to complete a right-handed orthogonal frame.}
            \\
		{\textit{body}-frame} & {The coordinates system of the Wheel-IMU.} & {\textit{origin}: IMU measurement center.\newline
				\textit{x-axis}: right, parallel to the rotation axis of the wheel.\newline
				 \textit{y}- and \textit{z-axes} are parallel to the wheel plane to complete a right-handed orthogonal frame.}
		\\
		\bottomrule
	\end{tabular}
\end{table}

\subsection{Wheel-INS}

Wheel-GINS, proposed in this paper, builds on top of our prior work Wheel-INS~\cite{niu2021}, a 2D dead reckoning system using only one wheel-mounted IMU. To be specific, we perform the strapdown INS to predict the vehicle state in Wheel-INS. At the same time, we multiply the gyroscope readings in the \textit{x}-axis of the Wheel-IMU with the wheel radius to calculate the vehicle forward velocity. After that, this velocity is integrated with the non-holonomic constraints as a full 3D velocity observation to correct the INS-indicated robot state via an EKF. We use the error-state formulation to mitigate the nonlinearity issue, which includes the vehicle position, velocity, and attitude errors, as well as the bias and scale factor error of the IMU sensor. We calibrate the Wheel-IMU leverarm and mounting angle offline in Wheel-INS.

Because the Wheel-IMU rotates with the wheel, Wheel-INS cannot accurately estimate the vehicle pitch angle. In other words, it cannot determine whether the vehicle is ascending or descending. Consequently, Wheel-INS assumes the vehicle moves on a horizontal plane, limiting its localization capabilities to two dimensions. 

\section{Our Approach}
Wheel-GINS is built on top of Wheel-INS. In Wheel-GINS, we fuse the absolute position information from GNSS with Wheel-INS using an EKF. As in the case of Wheel-INS, we use the error state as the system model to mitigate the nonlinearity issue. In addition to the vehicle velocity observation used in Wheel-INS, we construct the GNSS position observation model to limit the long-term error drift and the wheel angular velocity constraint model for online mounting angle estimation in Wheel-GINS. In this section, we elaborate on the error state model and the observation models used in Wheel-GINS.  

\subsection{Error State Model}
In Wheel-INS, we simplify the INS propagation model by omitting less significant terms, e.g., earth rotation, as it is only a local dead reckoning system using a low-cost inertial sensor. This simplification reduces computational overhead without compromising accuracy. In contrast, Wheel-GINS employs a full strapdown INS in the \textit{earth}-frame to integrate precise global position data from GNSS. For detailed information on the strapdown INS model, refer to~\cite{savage2007strapdown, Shin2005}. 

We adopt an error-state model in EKF to mitigate nonlinear errors in Wheel-GINS. In addition to the vehicle state and the IMU sensor errors, we augment the Wheel-IMU installation parameter errors (including the Wheel-IMU leverarm and mounting angle and the wheel radius scale error) into the error state vector, which can be written as
\begin{equation}
\bm{x} \!=\!\left[\delta\bm{p}^{\top}\!\quad\!\delta \bm{v}^{\top}\!\quad\!\bm{\phi}^{\top}\!\quad\!\bm{b}_{g}^{\top}\!\quad\!\bm{b}_{a}^{\top}\!\quad\!\bm{s}_{g}^{\top}\!\quad\!\bm{s}_{a}^{\top}\!\quad\!\delta\bm{l}_w^{\top}\!\quad\!\delta\bm{\phi}_m^{\top}\!\quad\!{s_r}\right]^{\top},
\end{equation}
where $\delta$ indicates the error of the vehicle state, particularly, $\delta \bm{p}$, $\delta \bm{v}$ and $\bm{\phi}\!\in\!\mathfrak{so}(3)$ indicate the position, velocity, and attitude errors of INS in the \textit{n}-frame, respectively; $\bm{b}_{g}$ and $\bm{b}_{a}$ represent the residual bias errors of the gyroscope and the accelerometer, respectively; $\bm{s}_{g}$ and $\bm{s}_{a}$ represent the residual scale factor errors of the gyroscope and accelerometer, respectively; $\delta\bm{l}_w\!\in\!\mathbb{R}^{2}$ is the residual Wheel-IMU leverarm error; $\delta\bm{\phi}_m\!\in\!\mathbb{R}^{2}$ is the residual Wheel-IMU mounting angle error; $s_r\!\in\!\mathbb{R}$ is the scale error of the wheel radius. Therefore, $\bm{x}$ is a vector with 26 dimensions. The IMU sensor errors are modeled using the first-order Gauss-Markov process, while the Wheel-IMU leverarm error, mounting angle error, and wheel radius scale error are modeled using random walks. 

We use the detailed error-state model~\cite{Shin2005,Groves2013} in Wheel-GINS instead of the simplified version used in Wheel-INS. Because Wheel-GINS aims to provide accurate positioning results in large-scale (kilometer-level) environments, instead of medium-scale (hundred-meters level) environments for Wheel-INS, we have to take more terms into consideration, for example, the earth rotation, the rotation rate of the \textit{n}-frame with respect to the \textit{earth}-frame and the change of gravity. The error state model is given by

\begin{equation}
	\left\{
	\begin{aligned}
	\delta \dot{\bm{p}}&= -\bm{\omega}^n_{en}\times\delta\bm{p} + \delta\bm{\phi}_{en} \times \bm{v} + \delta\bm{v} \\
	\delta \dot{\bm{v}}&= \mathbf{R}_b^n\bm{f}^{b} \times \bm{\phi} + \mathbf{R}_{b}^{n}\delta\bm{f}^{b} + \bm{v} \times \left(2\delta\bm{\omega}^n_{ie} + \delta\bm{\omega}^n_{en}\right) \\ &\quad - \left(2\bm{\omega}^n_{ie} + \bm{\omega}^n_{en}\right) \times \delta\bm{v} + \delta\bm{g}^n \\
	\dot{\bm{\phi}}&=-\bm{\omega}^n_{in} \times \bm{\phi} + \delta\bm{\omega}^n_{in} -\mathbf{R}_{b}^{n}\delta\bm{\omega}\\
	\dot{\bm{b}}_{g}&=-(1/\tau_{bg}) \bm{b}_{g}+\bm{w}_{b_g}\\
	\dot{\bm{b}}_{a}&=-(1/\tau_{ba}) \bm{b}_{a}+\bm{w}_{b_a}\\
	\dot{\bm{s}}_{g}&=-(1/\tau_{sg}) \bm{s}_{g}+\bm{w}_{s_g}\\
	\dot{\bm{s}}_{a}&=-(1/\tau_{sa}) \bm{s}_{a}+\bm{w}_{s_a}\\
	\dot{\delta\bm{l}_w}&=\bm{w}_{l_w}\\
	\dot{\delta\bm{\phi}_m}&=\bm{w}_{\phi_m}\\
	\dot{s}_r&=w_r
	\end{aligned},
	\right.
\end{equation}
where $\delta\bm{\phi}_{en}$ is the rotation vector describing the error of the INS-indicated \textit{n}-frame; $\bm{\omega}^n_{ie}$ is the earth's rotation rate described in the \textit{n}-frame, while $\delta\bm{\omega}^n_{ie}$ is its error; $\bm{\omega}^n_{en}$ is the rotation rate of the \textit{n}-frame with respect to the \textit{earth}-frame, while $\delta\bm{\omega}^n_{en}$ is its error; $\bm{\omega}^n_{in}$ is the rotation rate of the \textit{n}-frame, while $\delta\bm{\omega}^n_{in}$ is its error; $\mathbf{R}_b^n$ is the rotation matrix from the \textit{body}-frame to the \textit{n}-frame; $\tau_{b_g}$, $\tau_{b_a}$, $\tau_{s_g}$, and $\tau_{s_a}$ are the correlation time in the first-order Gauss-Markov model of the gyroscope bias, accelerometer bias, gyroscope scale factor, and accelerometer scale factor, respectively; $\bm{w}_{b_g}$ and $\bm{w}_{b_a}$ denote the driving white noise of the residual bias errors of the gyroscope and accelerometer, respectively; $\bm{w}_{s_g}$ and $\bm{w}_{s_a}$ denote the driving white noise of the scale factor errors of the gyroscope and accelerometer, respectively; $\bm{w}_{l_w}$, $\bm{w}_{\phi_m}$ and $w_r$ denote the driving white noise of the leverarm error, mounting angle error, and wheel radius scale error, respectively. More details of the error-state model can be found in Shin~\cite{Shin2005}.

\subsection{Observation Model}
In this section, we present a detailed derivation of the observation models, including the vehicle velocity observation, the GNSS position observation, and the wheel angular velocity observation used in the EKF scheme of the proposed Wheel-GINS. Different from Wheel-INS, we incorporate the Wheel-IMU installation parameters into the state vector for online estimation. Note that the Wheel-IMU leverarm and mounting angle errors are modeled as 2D vectors, while the wheel radius scale error is modeled as a scalar (see Section III-A). Although our derivation considers the full 3D vector of the Wheel-IMU leverarm and mounting angle, we extract only the elements relevant to the state vector for matrix construction in practical computation. 
\subsubsection{Vehicle Velocity Observation}
To compute the wheel velocity, we first need to know the rotation speed of the wheel. Therefore, we need to transform the Wheel-IMU angular velocity from the \textit{body}-frame to the \textit{wheel}-frame, namely,
\begin{equation}
\hat{\bm{\omega}}^w = \hat{\mathbf{R}}_{b}^{w}\hat{\bm{\omega}},
\end{equation}
where $\hat{\bm{\omega}}^w$ is the estimated angular velocity of the wheel in the \textit{wheel}-frame; $\hat{\mathbf{R}}_{b}^{w}$ is the rotation matrix from the \textit{body}-frame to the \textit{wheel}-frame; $\bm{\omega}$ is the angular velocity measurement of the Wheel-IMU in the \textit{body}-frame. Because we only need the wheel angular velocity in the \textit{x}-axis of the \textit{wheel}-frame to calculate the wheel velocity, we can simplify the equation as
\begin{equation}
\hat{\omega}_{x}^{w} = \hat{\mathbf{R}}_{b(1,:)}^{w}\hat{\bm{\omega}},
\end{equation}
where $\hat{\omega}_{x}^{w}$ is the estimated angular velocity of the wheel in the \textit{x}-axis of the \textit{wheel}-frame; $\hat{\mathbf{R}}_{b(1,:)}^{w}$ is the first row of the rotation matrix from the \textit{body}-frame to the \textit{wheel}-frame, where $\hat{\mathbf{R}}_{b(1,:)}^{w} = {\mathbf{R}}_{b(1,:)}^w + \delta{\mathbf{R}}_{b(1,:)}^w$, and $\delta{\mathbf{R}}_{b(1,:)}^w$ is the error of the first row of the rotation matrix, which is governed by the Wheel-IMU mounting angle error $\delta\bm{\phi}_m$.

Further, taking into consideration the wheel radius scale error, we can calculate the forward wheel velocity and derive the error as
\begin{equation}
\begin{aligned}
\widetilde{v}^{v}_{wheel} &=\hat{\omega}_{x}^{w}\hat{r}-e_v = \hat{\mathbf{R}}_{b(1,:)}^w\hat{\bm{\omega}}\hat{r}-e_v \\ 
&=({\mathbf{R}}_{b(1,:)}^w + \delta{\mathbf{R}}_{b(1,:)}^w)({\bm{\omega}}+\delta\bm{\omega})(1+s_r)r-e_v\\ 
&={v}^{v}_{wheel}+{\mathbf{R}}_{b(1,:)}^wr\delta\bm{\omega} + \mathbf{A}\delta\bm{\phi}_m + {\mathbf{R}}_{b(1,:)}^w\bm{\omega}{r}s_r- e_v,
\end{aligned}
\label{vehicle_vel_obs_model}
\end{equation}
where $\widetilde{v}^{v}_{wheel}$ and ${v}^{v}_{wheel}$ are the observed and true wheel velocity, respectively; $\delta\bm{\omega}$ is the error of the angular velocity measurement, which is $\delta\bm{\omega} = \bm{b}_{g} + \mathrm{diag}(\bm{\omega})\bm{s}_{g} + \bm{e}_\omega$, where $\bm{e}_\omega$ is the gyroscope noise and $\mathrm{diag}(\cdot)$ is the diagonal matrix form of a vector; $r$ is the wheel radius; $s_r$ is the wheel radius scale error, and $e_v$ is the observation noise, modeled as Gaussian white noise; $\mathbf{A}$ is the coefficent matrix of the mounting angle error, with dimensions $1\!\times\!2$. Please refer to the Appendix for the derivation of $\mathbf{A}$.

By combining the forward wheel velocity with the non-holonomic constraints, we can formulate the complete 3D vehicle velocity model as
\begin{equation}
\widetilde{\bm{v}}^{v}_{wheel} =\begin{bmatrix}
\widetilde{v}^{v}_{wheel} &\! \!0 &\!0
\end{bmatrix}^\top-\bm{e}_v.
\end{equation}

Simultaneously, through perturbation analysis, the wheel velocity in the \textit{vehicle}-frame indicated by the INS can be expressed as
\begin{equation}
\begin{aligned}
\hat{\bm{v}}^v_{wheel} &= \hat{\mathbf{R}}_n^v \hat{\bm{v}}\!+\!\hat{\mathbf{R}}^v_b \left( \hat{\bm{\omega}} \times \right) \hat{\bm{l}}_{w} \\
&\approx {\mathbf{R}}^v_n (\mathbf{I} \!+\! \delta\bm{\psi} \times)({\bm{v}} \!+\! \delta{\bm{v}}) \\ 
&\quad \!+\! (\mathbf{I} \!-\! \delta\bm{\phi}_m \times) \mathbf{R}^v_b ({\bm{\omega}}\!\times \!+\! \delta{\bm{\omega}} \times)(\bm{l}_{w} \!+\! \delta{\bm{l}_{w}} )\\
&\approx {\bm{v}}^v_{wheel} \!+\! {\mathbf{R}}_n^v \delta{\bm{v}}^n \!-\! {\mathbf{R}}^v_n ({\bm{v}} \times) \delta\bm{\psi} \!-\! {\mathbf{R}}^v_b(\bm{l}_{w} \times)\delta{\bm{\omega}}\\
&\quad \!+\! {\mathbf{R}}^v_b (\bm{\omega}\times)\delta{\bm{l}_{w}} \!+\! \left({\mathbf{R}}^v_b (\bm{\omega}\times){\bm{l}_{w}}\right)\!\times\!\delta\bm{\phi}_m,
\end{aligned}
\label{insvspeed}
\end{equation} 
where $\mathbf{R}_{n}^{v}$ is the rotation matrix from the \textit{n}-frame to the \textit{vehicle}-frame; $\mathbf{R}_{b}^{v}$ is the rotation matrix from the \textit{body}-frame to the \textit{vehicle}-frame; $\bm{\omega}$ is the angular velocity measurement of the Wheel-IMU; $(\cdot)\times$ indicates the skewsymmetric matrix of a vector; $\bm{l}_{w}$ indicates the leverarm vector from the Wheel-IMU to the wheel center expressed in the \textit{body}-frame, while $\delta\bm{l}_{w}$ indicates its residual error; $\delta\bm{\psi}$ is the attitude error of the vehicle. Because Wheel-INS cannot estimate the vehicle pitch angle, it is assumed that the vehicle is moving on a horizontal plane in Wheel-INS, which means the vehicle attitude only includes heading angle as the other two components are zero. Therefore, the vehicle attitude error is only related to the heading error of the attitude error in the state vector, which can be written as $\delta\bm{\psi} = \begin{bmatrix}
0 &\! 0 &\! \delta{\psi}
\end{bmatrix}^\top$.

Then, the vehicle velocity measurement model can be expressed as
\begin{equation}
\begin{aligned}
\delta\bm{z}_v &= \hat{\bm{v}}^v_{wheel} - \widetilde{\bm{v}}^v_{wheel}\\
&= \mathbf{H}_v\bm{x}+\bm{e}_v,
\end{aligned}
\end{equation}
where $\mathbf{H}_v$ is a $3\!\times\!26$ matrix of the form
\begin{equation}
\mathbf{H}_v \!=\! 
\begin{bmatrix}
\mathbf{0} &\mkern-8mu {\mathbf{R}}_n^v &\mkern-8mu -{\mathbf{R}}^v_n ({\bm{v}} \times) &\mkern-8mu -{\mathbf{R}}^v_b(\bm{l}_{w} \times) &\mkern-8mu \mathbf{0} &\mkern-8mu \mathbf{B} &\mkern-8mu \mathbf{0} &\mkern-8mu \mathbf{C} &\mkern-8mu \mathbf{D} &\mkern-8mu \mathbf{E}
\end{bmatrix},
\end{equation}
and 
\begin{equation}
\left\{
\begin{aligned}
\mathbf{B} &= -{\mathbf{R}}^v_b(\bm{l}_{w} \times)\mathrm{diag}(\bm{\omega}) \\ 
\mathbf{C} &= {\mathbf{R}}^v_b (\bm{\omega}\times)_{(:,2:3)} \\ 
\mathbf{D} &= \left({\mathbf{R}}^v_b (\bm{\omega}\times){\bm{l}_{w}}\right)\!\times_{(:,2:3)} -\left[\mathbf{A}\quad \mathbf{0} \quad \mathbf{0} \right]^{\top} \\ 
\mathbf{E} &= \left[-{\mathbf{R}}_{b(1,:)}^w\bm{\omega}{r} \quad {0} \quad {0}\right]^{\top}
\end{aligned}
\right.
\label{systemnodel}
\end{equation}
where $_{(:,2:3)}$ indicates the second and third columns of a matrix. Note that we take only the last two columns of the matrix corresponding to the actual misalignment error to build $\mathbf{C}$ and $\mathbf{D}$ in $\mathbf{H}_v$ because the Wheel-IMU leverarm error and mounting angle error are in 2-dimension as can be seen in Eq.~2 and Eq.~3.

In addition, one may argue that the observed wheel velocity here is computed with the gyroscope measurement. This means that the observation is correlated with the system state, which does not conform to the prerequisite of the EKF. However, in our previous Wheel-INS paper \cite{niu2021}, we have illustrated that this issue does not significantly affect the performance of the system. Wheel-INS can still provide accurate and reliable localization results. Therefore, we also do not take any specific measures to address this issue in this study. 

\subsubsection{GNSS Observation}
 
We calculate the difference between the observed GNSS antenna position from the receiver and the INS-predicted antenna position to build the GNSS observation model. Given the IMU position propagated by the strapdown INS, the GNSS antenna's position in the \textit{earth}-frame can be derived as
\begin{equation}
    \hat{\bm{p}}^e_{gnss} = \hat{\bm{p}}^e_{b} +\mathbf{R}_n^e\hat{\mathbf{R}}_b^n\hat{\bm{l}}_{gnss},
\label{imu_predicted_antenna_pos}
\end{equation}
where $\mathbf{R}_n^e$ is the rotation matrix from the local navigation frame to the \textit{earth}-frame; $\bm{l}_{gnss}$ is the GNSS leverarm, equaling to the vector from the Wheel-IMU center to the GNSS antenna center expressed in the \textit{body}-frame.

Unlike the traditional GNSS/INS integrated navigation system where the leverarm between the GNSS antenna and the IMU is fixed, the GNSS leverarm expressed in the \textit{body}-frame changes in Wheel-GINS because the Wheel-IMU continuously rotates with the wheel. Therefore, the GNSS leverarm has to be computed in real-time in Wheel-GINS. 

Given that the origin of the \textit{vehicle}-frame is aligned with the wheel center, the position of the GNSS antenna is constant in the \textit{vehicle}-frame, which can be measured beforehand. Then, the GNSS leverarm w.r.t the \textit{body}-frame can be calculated as
\begin{equation}
\hat{\bm{l}}_{gnss} = \hat{\bm{l}}_w + \hat{\mathbf{R}}_v^b\bm{l}_{gnss}^v.
\end{equation}

Consequently, we can get the GNSS observation equation
\begin{equation}
\begin{aligned}
    \delta{\bm{z}}_{gnss} &=\mathbf{R}_e^n\left(\hat{\bm{p}}^e_{gnss}-\tilde{\bm{p}}^e_{gnss}\right)\\
    &=\delta\bm{p} + \left(\mathbf{R}_b^n\left(\bm{l}_w + \mathbf{R}_v^b\bm{l}_{gnss}^v\right)\right) \!\times\! \bm{\phi} \\
    &\quad -\mathbf{R}_b^n\mathbf{R}_v^b\left(\bm{l}_{gnss}^v\times \right) \delta\bm{\phi}_m + \mathbf{R}_b^n \delta\bm{l}_{w}\\
    &= \mathbf{H}_{gnss}\bm{x} +\bm{e}_{gnss},
\end{aligned}
\end{equation}
where $\mathbf{H}_{gnss}$ is a $3\!\times\!26$ matrix of the form
\begin{equation}
\begin{aligned}
\mathbf{H}_{gnss} &= 
\begin{bmatrix}
\mathbf{I} &\mkern-8mu \mathbf{0} &\mkern-8mu \mathbf{F} &\mkern-8mu \mathbf{0} &\mkern-8mu \mathbf{0} &\mkern-8mu \mathbf{0} &\mkern-8mu \mathbf{0} &\mkern-8mu  {\mathbf{R}}^n_{b(:,2:3)} &\mkern-8mu \mathbf{G} &\mkern-8mu \mathbf{0}
\end{bmatrix},
\end{aligned}
\end{equation}
and
\begin{equation}
	\left\{
	\begin{aligned}
	\mathbf{F} &= \left(\mathbf{R}_b^n\left(\bm{l}_w + \mathbf{R}_v^b\bm{l}_{gnss}^v\right)\right) \!\times\! \\
	\mathbf{G} &= -\mathbf{R}_b^n\mathbf{R}_v^b\left(\bm{l}_{gnss}^v\times \right)_{(:,2:3)}
	\end{aligned}.
	\right.
\end{equation}

\subsubsection{Wheel Angular Velocity Observation}
To effectively estimate the Wheel-IMU mounting angle online, we construct and integrate a wheel angular velocity observation model into Wheel-GINS. Note that the wheel angular velocity constraint is only used for the Wheel-IMU mounting angle estimation, not to help with the robot state estimation. 

Assuming that the vehicle travels in a straight line without turning and ignoring the Earth's rotation, the wheel only has angular velocity along the rotation axis, namely, 
\begin{equation}
\widetilde{\bm{\omega}}^{w} =\begin{bmatrix}
\widetilde{\omega}^{w}_{x} &\! \!0 &\!0
\end{bmatrix}^\top-\bm{e}_{\omega},
\label{wheel_angular_velocity}
\end{equation}
where $\widetilde{\bm{\omega}}^{w}$ indicates the angular velocity of the wheel in the \textit{wheel}-frame; $\widetilde{{\omega}}_x^{w}$ is the \textit{x}-axis component of $\widetilde{\bm{\omega}}^{w}$; $\bm{e}_{\omega}$ indicates the measurement noise. As a result, the angular velocity sensed by the \textit{y}-axis and \textit{z}-axis of the Wheel-IMU should equal zero if there is no attitude misalignment error. Therefore, we can construct an observation model based on this fact to estimate the attitude misalignment between the Wheel-IMU and the wheel. The angular velocity of the wheel in the \textit{wheel}-frame computed with the Wheel-IMU gyroscope readings can be expressed as
\begin{equation}
\begin{aligned}
    \hat{\bm{\omega}}^w &= \hat{\mathbf{R}}_b^w\hat{\bm{\omega}} \\
    &= (\mathbf{I} - \bm{\phi}_{m}\times){\mathbf{R}}_b^w({\bm{\omega}} + \delta{{\bm{\omega}}}),
\end{aligned}
\end{equation}
where ${\mathbf{R}}_b^w$ indicates the rotation matrix from the \textit{body}-frame to the \textit{wheel}-frame. Then, the wheel angular velocity observation model can be written as  
\begin{equation}
\begin{aligned}
    \delta{\bm{z}_\omega} &= \hat{\bm{\omega}}^w - \tilde{\bm{\omega}}^w \\
    &= {\mathbf{R}}_b^w\delta{{\bm{\omega}}} + ({\mathbf{R}}_b^w{\bm{\omega}})\!\times\!\delta\bm{\phi}_{m}\\
    &= \mathbf{H}_{\omega}\bm{x} +\bm{e}_{\omega},
\label{angular_velocity_observation}
\end{aligned}
\end{equation}
where $\mathbf{H}_{\omega}$ is a $2\!\times\!26$ matrix of the form
\begin{equation}
\mathbf{H}_{\omega} =
\begin{bmatrix}
\mathbf{0} &\mkern-8mu \mathbf{0} &\mkern-8mu \mathbf{0} &\mkern-8mu {\mathbf{R}}_b^w &\mkern-8mu \mathbf{0} &\mkern-8mu \mathbf{R}_b^w\mathrm{diag}(\bm{\omega}) &\mkern-8mu \mathbf{0} &\mkern-8mu  \mathbf{0} &\mkern-8mu ({\mathbf{R}}_b^w{\bm{\omega}})\!\times_{(:,2:3)} &\mkern-8mu \mathbf{0}
\end{bmatrix}.
\end{equation}

Note that we use only the bottom two rows of the matrices on both sides of Eq.~19 to construct the observation model because only the \textit{y}-axis and \textit{z}-axis component of the wheel angular velocity in the \textit{wheel}-frame should equal zero. Therefore, we take only the last two rows of $\mathbf{H}_{\omega}$.

One may argue that the assumption of the vehicle moving along a straight line is strong because the vehicle inevitably turns during motion. The turning of the vehicle will project angular velocity to the \textit{y}-axis and \textit{z}-axis of the wheel, making Eq.~\ref{wheel_angular_velocity} not strictly valid. However, in practice, we can detect the vehicle turning with the Wheel-IMU. Thus, we can employ the wheel angular velocity observation model to estimate the Wheel-IMU mounting angle specifically when the vehicle is moving straight and subsequently fix it once convergence is achieved. Our experimental results have shown that the Wheel-IMU mounting angle converges fast. In addition, the occasional normal turning of the vehicle has negligible impact on both the Wheel-IMU mounting angle and vehicle pose estimation (see Section V-C-2).

\begin{table}[t]
	\centering
	\caption{Technical Parameters of the IMUs Used in the Tests}
	\label{tab:IMU_parameters}
	\begin{threeparttable}
		\begin{tabular}{p{3cm}<{\centering}p{1.2cm}<{\centering}p{1.2cm}<{\centering}p{1.2cm}<{\centering}}
			\toprule
			IMU & iNav-FJI & POS320 & ICM20602\\
			\midrule
			{Gyro Bias ($deg/h$)}            & 0.01    & 0.5     & 200\\
			{ARW\tnote{*}  ($deg/\sqrt{h}$)} & 0.001   & 0.05    & 0.24\\
			{Acc\tnote{*} Bias  ($m/s^2$)}   & 0.00016 & 0.00025 & 0.01\\
			{VRW\tnote{*}  ($m/s/\sqrt{h}$)} & 0.0009  & 0.1     & 3\\
			{Gyro scale factor std\tnote{*}} & 0.00003 & 0.001   & 0.03\\
			{Acc scale factor std}           & 0.00016 & 0.001   & 0.03\\
			\bottomrule
		\end{tabular}
		\begin{tablenotes}   
			\footnotesize            
			\item[*]ARW denotes the angle random walk; Acc denotes accelerometer; VRW denotes velocity random walk; std denotes standard deviation.      
		\end{tablenotes}            
	\end{threeparttable}
\end{table}

\begin{table}[t]
	\centering
	\caption{Vehicle Motion Information in the Experiments}
	\label{tab:vehicle_motion_info}
	\begin{tabular}{cccc}
		\toprule
 		 Sequence & Vehicle & \makecell{Average \\ speed ($m/s$)} & \makecell{Total \\ distance ($m$)} \\
		\midrule
		\specialrule{0em}{2pt}{2pt}
		  {1} & {Pioneer 3DX}  & \makecell{1.25 } & \makecell{$\approx$1146 }\\
            \specialrule{0em}{2pt}{2pt}
            {2} & {Trolley}  & \makecell{1.41 } & \makecell{$\approx$1990 }\\
            \specialrule{0em}{2pt}{2pt}
            {3} & {Car}  & \makecell{6.00 } & \makecell{$\approx$6566 }\\
		\bottomrule
	\end{tabular}   
\end{table}

\section{Experimental Results}
This paper presents Wheel-GINS, a GNSS/INS integrated navigation system with a wheel-mounted IMU. As discussed in the introduction, Wheel-GINS achieves information fusion in a similar way to ODO-GINS. Therefore, we used ODO-GINS as the benchmark to illustrate the performance of the proposed Wheel-GINS. This section presents real-world experimental results and analysis to support our key claims, namely, (i) Wheel-GINS has significantly reduced the position error drift compared to ODO-GINS during GNSS outages; (ii) Wheel-GINS can effectively estimate the Wheel-IMU installation parameters, including the Wheel-IMU leverarm and mounting angle and the wheel radius scale error online, thus improving the pose estimation accuracy.
\begin{figure}[t]
	\centering
	\includegraphics[width=8.8cm]{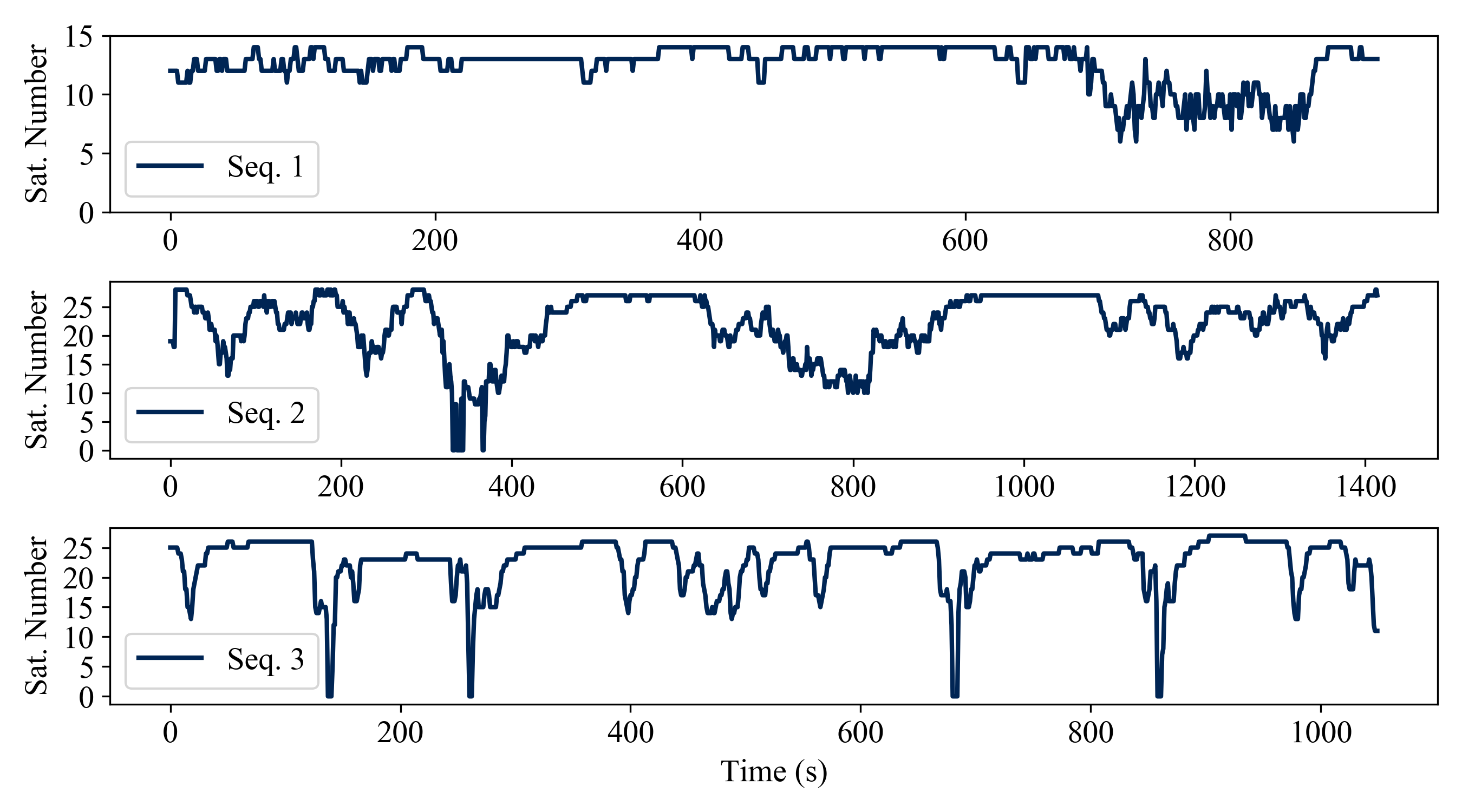}
	\caption{The number of GNSS satellites used in the three sequences.}
	\label{fig:sat_number}
\end{figure}

\begin{figure*}[t]
	\centering
	\subfigure[Experimental robot and the devices used in Seq.~1.]{
		\includegraphics[width=8.5cm]{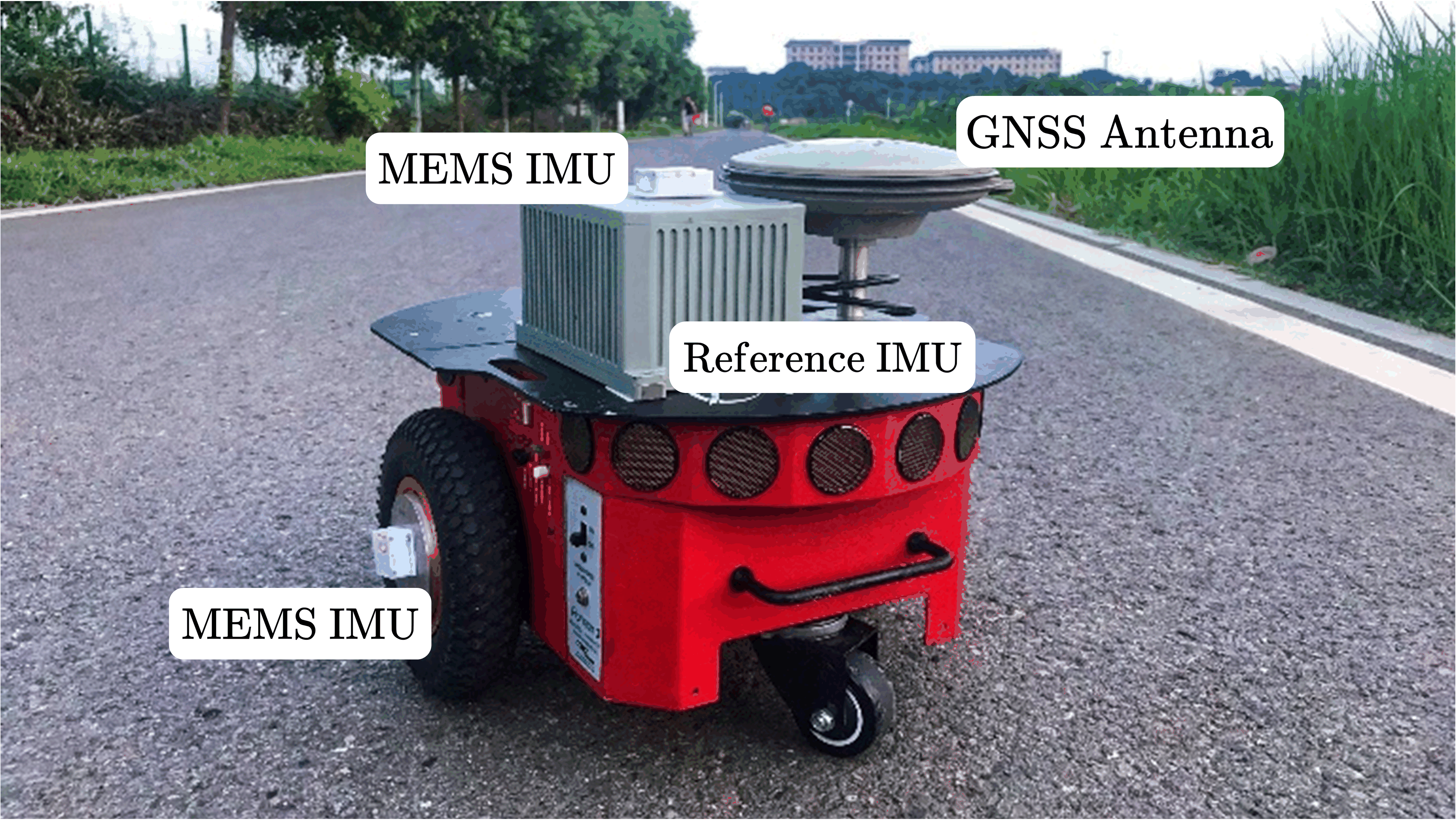}
	}
	\quad
	\subfigure[Seq.~1: polyline trajectory with no return.]{
		\includegraphics[width=8.5cm]{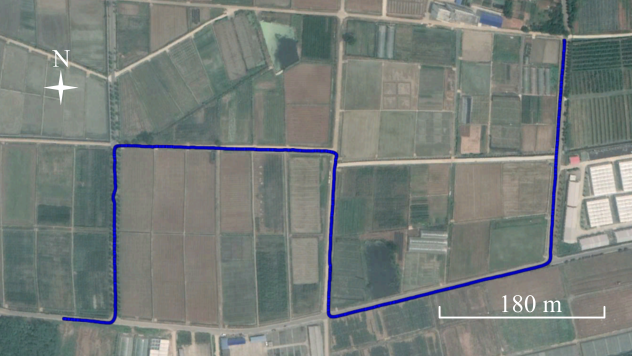}
	}
	\quad
	\subfigure[Experimental trolley and the devices used in Seq.~2.]{
		\includegraphics[width=8.5cm]{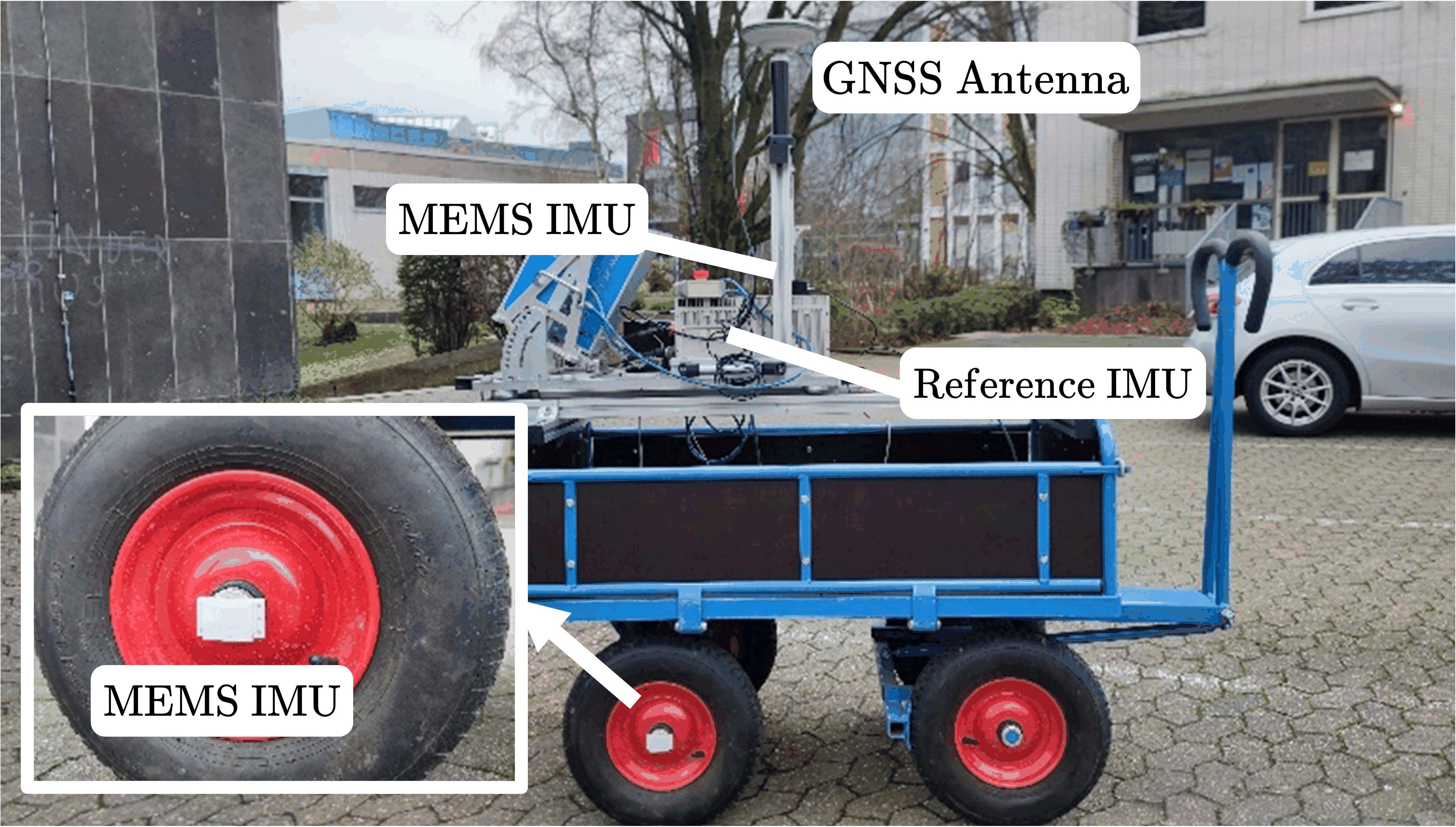}
	}
	\quad
	\subfigure[Seq.~2: loopback trajectory with overlap.]{
		\includegraphics[width=8.5cm]{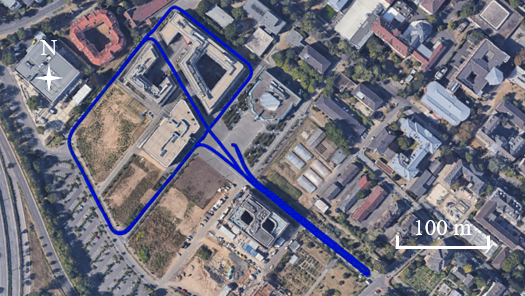}
	}
	\quad
	\subfigure[Experimental car and the devices used in Seq.~3.]{
		\includegraphics[width=8.5cm]{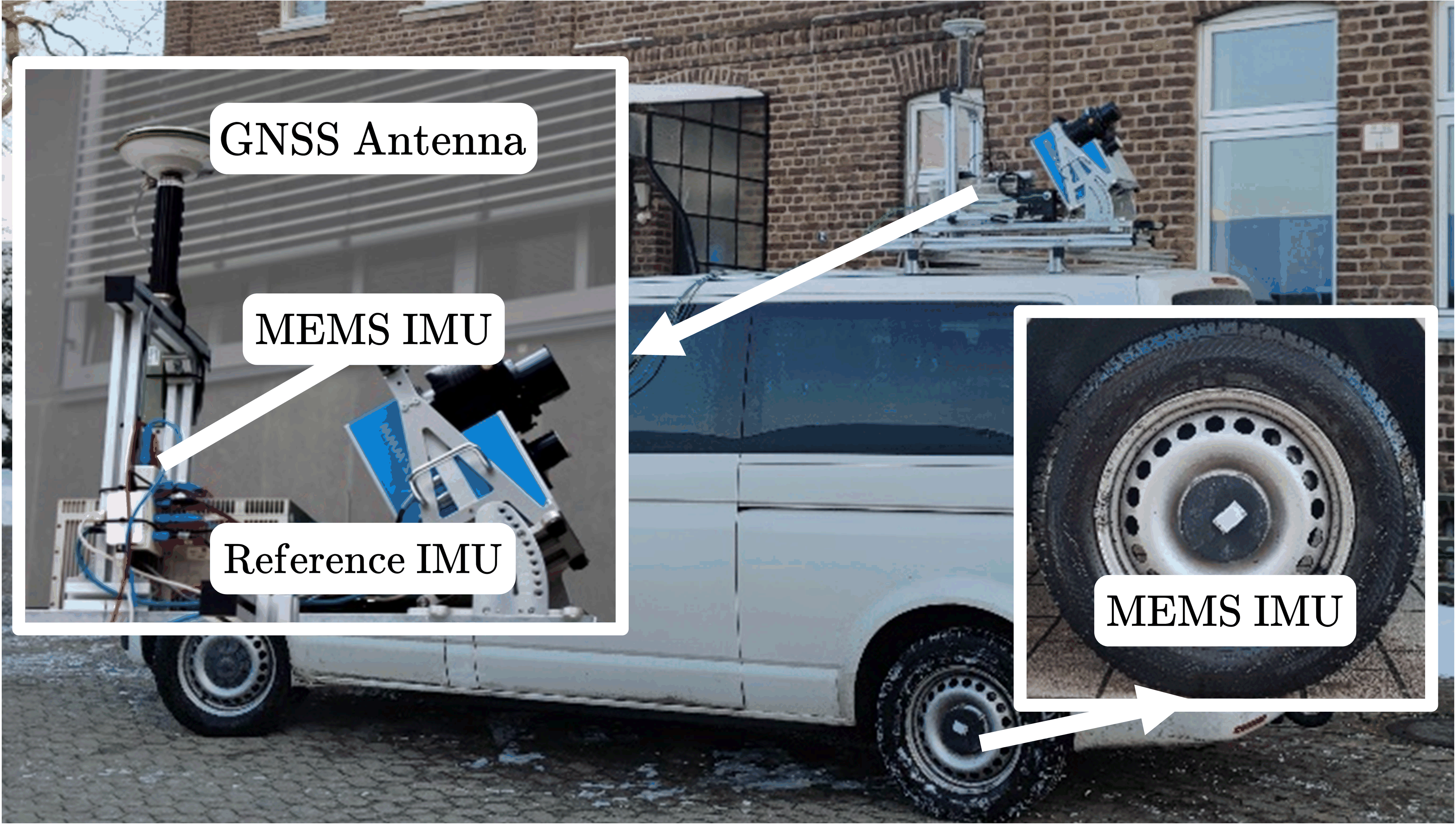}
	}
	\quad
	\subfigure[Seq.~3: large-scale trajectory with twice repetition.]{
		\includegraphics[width=8.5cm]{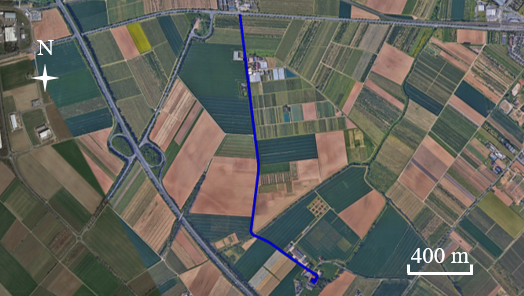}
	}
	\caption{The experimental platforms and trajectories in the three experimental sequences.}
	\label{fig:platforms_and_trajs}
\end{figure*}

\subsection{Experimental Setup}
To evaluate the performance of Wheel-GINS, we collected real-world data in three different places with three different wheeled vehicles. One was the Pioneer 3DX robot\footnote{\url{https://www.cyberbotics.com/doc/guide/pioneer-3dx?version=R2021a}}, the other was a trolley, and the third was a regular street car. Fig.~\ref{fig:platforms_and_trajs} shows the experimental platforms and the corresponding trajectories. The Wheel-IMU used for the experiments was self-developed. It contains a low-cost ICM20602\footnote{\url{https://invensense.tdk.com/download-pdf/icm-20602-datasheet/}} inertial sensor chip, a rechargeable battery, and a Bluetooth module for data transmission.
An IMU of the same model was mounted on the top of the vehicle to perform ODO-GINS for comparison. Because we didn't have an external odometer and it was difficult to access the wheel encoder of the vehicles, we calibrated the Wheel-IMU and compensated its error in advance to get the wheel velocity for ODO-GINS, but we didn't do this for Wheel-GINS. Therefore, the wheel velocity used in ODO-GINS is more accurate than that calculated in Wheel-GINS. 

As shown in Fig.~\ref{fig:platforms_and_trajs}, all three vehicles were equipped with a high-end IMU to provide a near ground truth trajectory: POS320\footnote{\url{http://www.whmpst.com/en/imgproduct.php?aid=32}}, a tactical-grade IMU for Sequence (Seq.) 1 and IMAR iNav-FJI\footnote{\url{https://www.imar-navigation.de/downloads/NAV_FJI_001-J_en.pdf}}, a navigation-grade IMU for Seq. 2 and Seq. 3. The main technical parameters of all IMUs are listed in Table \ref{tab:IMU_parameters}. We performed a smoothed post-processed kinematic (PPK)/INS integration method to compute this near ground truth trajectory. We used the same PPK GNSS position for both Wheel-GINS and ODO-GINS. The satellite numbers in the three sequences over time are shown in Fig.~\ref{fig:sat_number}.

As shown in Fig.~\ref{fig:platforms_and_trajs}, Seq. 1 is a one-way polyline trajectory with no return on an experimental field at a university; Seq.~2 is a loopback trajectory on a university campus; Seq.~3 is a large-scale loop trajectory where the vehicle drove twice on the same road. The average speed and the total traveled distance of the vehicles in the three sequences are listed in Table~\ref{tab:vehicle_motion_info}. 

We set the initial heading and position of both Wheel-GINS and ODO-GINS with the reference system. In real applications, the system initialization problem can be solved by online alignment approaches~\cite{chen2023ieeesj}. We calibrated the attitude misalignment between the reference IMU and the vehicle, as well as between the IMU mounted on the vehicle body and the vehicle in advance using the method proposed in Chen~\etalcite{chen2021}. Furthermore, we set the initial gyroscope bias using the static IMU data collected before the vehicle started moving. Other inertial sensor errors and the Wheel-IMU installation parameter errors were initialized as zero. The GNSS update frequency was set to \SI{1}{Hz} while the velocity update frequency was set to \SI{2}{Hz} for both Wheel-GINS and ODO-GINS.

\subsection{Comparison of Positioning Accuracy with ODO-GINS}
Table~\ref{tab:comparison} presents the position and heading root mean square error (RMSE) of the proposed Wheel-GINS and ODO-GINS in the three sequences. It can be observed from Table~\ref{tab:comparison} that Wheel-GINS achieves comparable accuracy to ODO-GINS: the positioning error is at the centimeter level, and the heading error is mostly less than \SI{1}{\degree}. In Seq. 1, the heading RMSE of Wheel-GINS is much larger than ODO-GINS. This is because the unevenness of the road caused significant vibration of the robot when it was moving. At the same time, the reference IMU was placed on top of the robot, not on the wheel. As a consequence, there is a larger error in the Wheel-GINS attitude estimates.

We can also see that with the integration of GNSS position observation, Wheel-GINS has improved Wheel-INS from a 2D dead reckoning system to a full 3D positioning system with accurate height estimation. In addition, Wheel-GINS does not show significant advantages compared to ODO-GINS when GNSS is always available. This is because the high-quality GNSS position observation has effectively suppressed the error drift of Wheel-INS and the traditional odometer-aided INS and helped estimate the IMU sensor errors.  

\begin{table}[t]
	\centering
	\caption{Pose Accuracy Statistics of Wheel-GINS and ODO-GINS}
	\label{tab:comparison}
	\begin{threeparttable}
		\begin{tabular}{p{0.2cm}<{\centering}p{1.5cm}<{\centering}p{1.5cm}<{\centering}p{1.4cm}<{\centering}p{1.4cm}<{\centering}}
			\toprule
			\multirow{2}*{\makecell{Seq.}} & \multirow{2}*{\makecell{System}} & {Horizon pos. RMSE (m)} &  {Height RMSE (m)} & {Heading RMSE (\SI{}{\degree})} \\
			\specialrule{0em}{1pt}{1pt}
			\midrule
			\specialrule{0em}{1.5pt}{1.5pt}
			\multirow{2}{*}{1}&{Wheel-GINS}& $\bm{0.04}$& $\bm{0.03}$&$1.15$\\
                \specialrule{0em}{1pt}{1pt}
			                     &{ODO-GINS}  & $0.05$& $0.05$&$\bm{0.61}$\\
			\specialrule{0em}{1.5pt}{1.5pt}
			\multirow{2}{*}{2}&{Wheel-GINS}& $\bm{0.07}$& $\bm{0.04}$&$\bm{0.39}$\\
                \specialrule{0em}{1pt}{1pt}
			                     &{ODO-GINS}  & $0.10$& $\bm{0.04}$&$\bm{0.39}$\\
			\specialrule{0em}{1.5pt}{1.5pt}
			\multirow{2}{*}{3}&{Wheel-GINS}& $0.10$& $0.26$&$\bm{0.35}$\\
                \specialrule{0em}{1pt}{1pt}
			                     &{ODO-GINS}  & $\bm{0.09}$& $\bm{0.13}$&$0.38$\\
			\bottomrule
		\end{tabular}
		\begin{tablenotes}   
			\footnotesize            
			\item[*]RMSE denotes root mean square error.      
		\end{tablenotes} 
	\end{threeparttable}
\end{table}

\begin{figure}[t]
	\centering
	\subfigure[Horizon position errors of Wheel-GINS and ODO-GINS with two \SI{30}{s} GNSS outages in Seq.~2.]{
		\includegraphics[width=8.8cm]{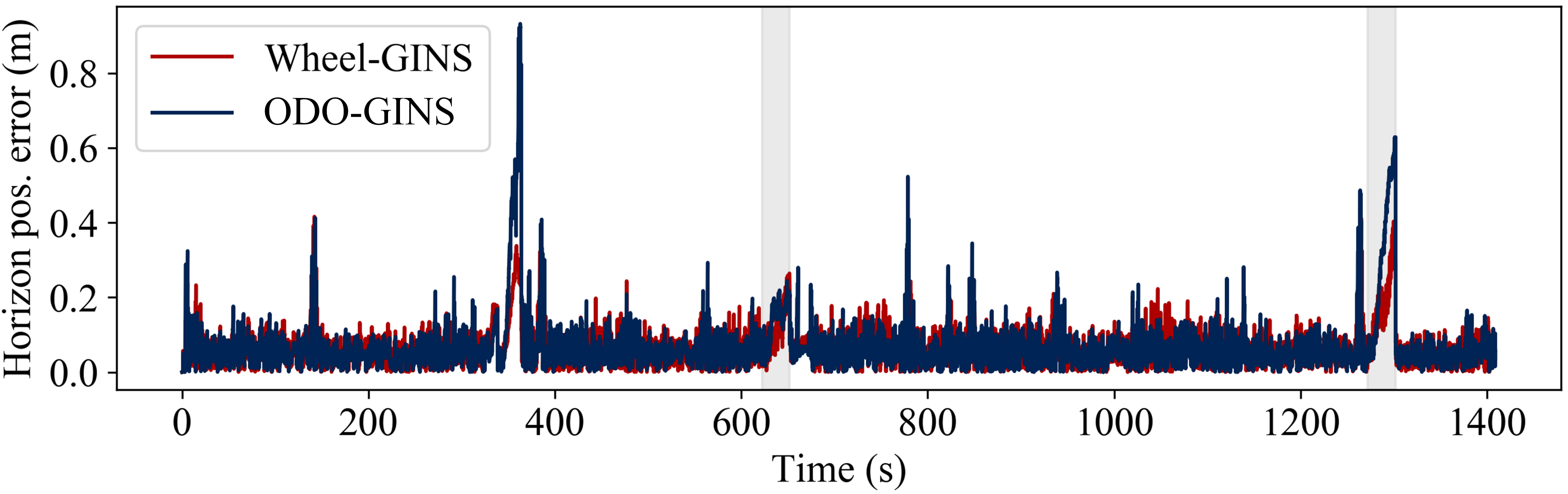}
	}
	\quad
	\subfigure[Horizon position errors of Wheel-GINS and ODO-GINS with two \SI{60}{s} GNSS outages in Seq.~2.]{
		\includegraphics[width=8.8cm]{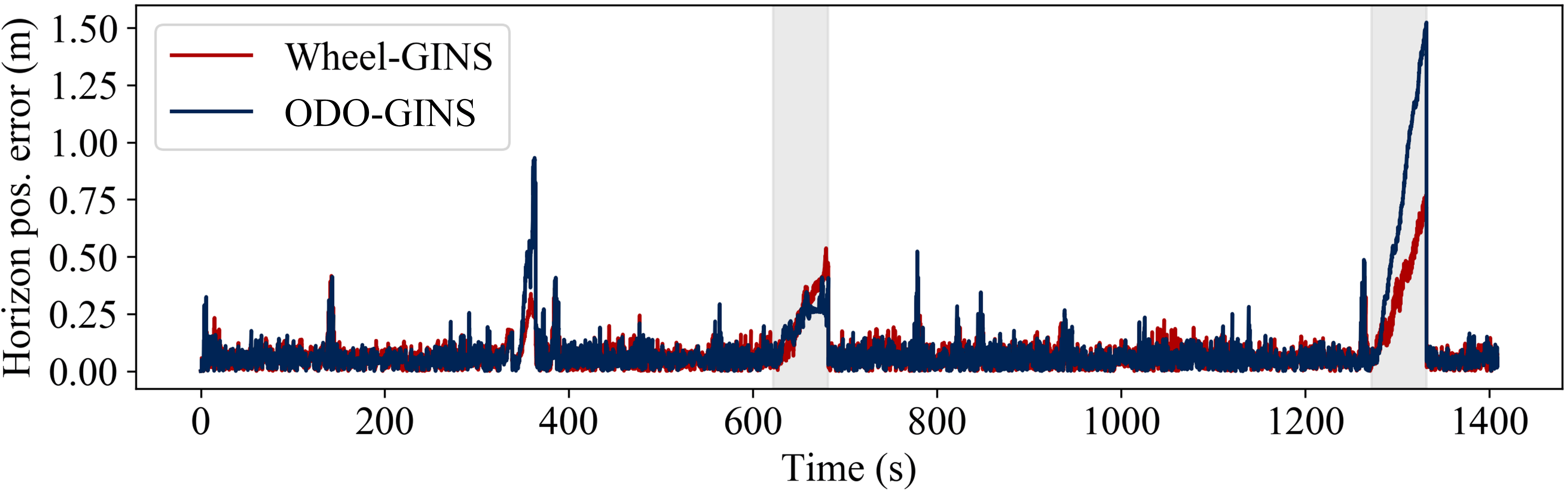}
	}
	\quad
	\subfigure[Horizon position errors of Wheel-GINS and ODO-GINS with two \SI{120}{s} GNSS outages in Seq.~2.]{
		\includegraphics[width=8.8cm]{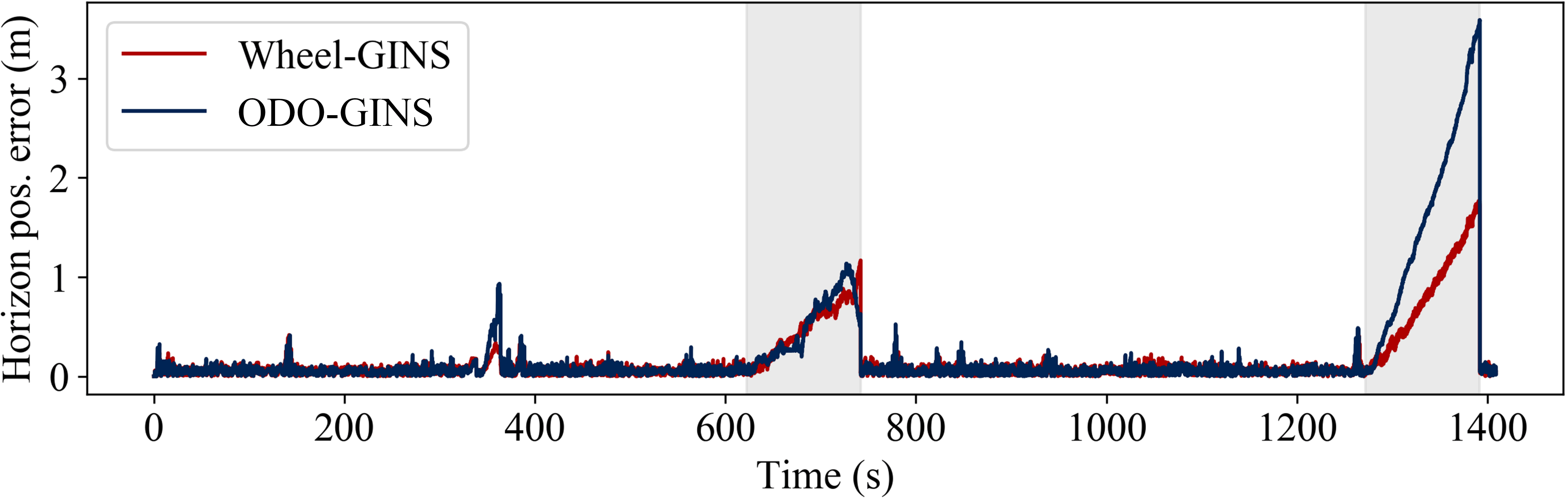}
	}
	\caption{Horizont position errors of Wheel-GINS and ODO-GINS with different lengths of GNSS outages in Seq.~2. The gray areas indicate the GNSS outage period.}
	\label{fig:position_err_outage_seq2}
\end{figure}

A commonly used method to evaluate the performance of a GNSS-aided integrated navigation system is to investigate the error drift when GNSS is blocked~\cite{niu2023tmech, chang2019rs,zhang2024sj,xiaozhu2018if}. To this end, we manually set three different lengths (namely, \SI{30}{s}, \SI{60}{s}, and \SI{120}{s}) of GNSS outages in the three sequences to compare the performance of Wheel-GINS with ODO-GINS. We set two outages for each sequence. Fig.~\ref{fig:position_err_outage_seq2} shows the horizon positioning error of Wheel-GINS and ODO-GINS with the three different lengths of GNSS outage in Seq. 2. Table~\ref{tab:comparison_outage} lists the mean RMSE and MAX horizontal position error of the two systems during GNSS outages in the three sequences.

We can see from Fig.~\ref{fig:position_err_outage_seq2} that without the global position information during the GNSS outages, the horizon positioning errors of both Wheel-GINS and ODO-GINS are growing. The horizon position error accumulation increases with the increase of the GNSS outage period. Due to the random error characteristics of the IMU, the drift rate is different at different points of the sequence with the same length of GNSS outage period. Note that the error drifts at around \SI{350}{s} in Fig.~\ref{fig:position_err_outage_seq2} are caused by the inferior GNSS conditions where the trolley was surrounded by high buildings.

In addition, we can see from Table~\ref{tab:comparison_outage} that Wheel-GINS exhibits higher accuracy than ODO-GINS during GNSS outages. Compared to ODO-GINS, the horizontal position RMSE of Wheel-GINS in \SI{30}{s}, \SI{60}{s}, and \SI{120}{s} GNSS outages has been averagely reduced by $28\%$, $32\%$, and $37\%$, respectively. The reason for this is that Wheel-INS exhibits a lower error drift rate than the traditional odometer-aided INS, as illustrated in~\cite{niu2021}. Even though GNSS helps to limit the error drift when it is available, Wheel-GINS outperforms ODO-GINS during GNSS outages thanks to the inherent rotation modulation effect. 

\begin{table}[t]
    \centering
    \caption{Comparison of Pose Estimation Error between Wheel-GINS and ODO-GINS during GNSS Outages}
    \label{tab:comparison_outage}
    \begin{threeparttable}
        \begin{tabular}{p{0.2cm}<{\centering}p{1.5cm}<{\centering}p{0.6cm}<{\centering}p{0.6cm}<{\centering}|p{0.6cm}<{\centering}p{0.6cm}<{\centering}|p{0.6cm}<{\centering}p{0.6cm}<{\centering}}
	\toprule
    \specialrule{0em}{1.5pt}{1.5pt}
	\multirow{2}*{\makecell{Seq.}} & \multirow{2}*{\makecell{System}} &        \multicolumn{6}{c}{Horizontal position ($m$)} \\
	\specialrule{0em}{1.5pt}{1.5pt}
	& & RMSE & MAX & RMSE & MAX & RMSE & MAX\\
        \midrule
        \specialrule{0em}{1.5pt}{1.5pt}
        \multicolumn{2}{c}{\centering\arraybackslash Outage time} & \multicolumn{2}{c|}{\centering\arraybackslash \SI{30}{s} } & \multicolumn{2}{c|}{\centering\arraybackslash \SI{60}{s}} & \multicolumn{2}{c}{\centering\arraybackslash \SI{120}{s}}\\
        \specialrule{0em}{1.5pt}{1.5pt}
        \midrule
        \multirow{2}{*}{1}&{Wheel-GINS}& $\bm{0.16}$& $\bm{0.27}$&$\bm{0.26}$&$\bm{0.62}$&$\bm{0.68}$&$\bm{1.28}$\\
        \specialrule{0em}{1pt}{1pt}
			             &{ODO-GINS}  & ${0.32}$& $0.49$&${0.44}$&$0.70$&${0.93}$&$1.86$\\
        \specialrule{0em}{1.5pt}{1.5pt}
        \multirow{2}{*}{2}&{Wheel-GINS}& $\bm{0.16}$& $\bm{0.33}$&$\bm{0.34}$&$\bm{0.65}$&$\bm{0.73}$&$\bm{1.47}$\\
        \specialrule{0em}{1pt}{1pt}
			             &{ODO-GINS}  & ${0.25}$& $0.44$&${0.53}$&$0.97$&${1.23}$&$2.36$\\
        \specialrule{0em}{1.5pt}{1.5pt}
        \multirow{2}{*}{3}&{Wheel-GINS}& $0.38$& $\bm{0.67}$&$\bm{0.75}$&$\bm{1.37}$&$\bm{1.37}$&$\bm{2.23}$\\
        \specialrule{0em}{1pt}{1pt}
			             &{ODO-GINS}  & $\bm{0.37}$& $0.71$&${0.92}$&$1.91$&${2.45}$&$4.46$\\
        \specialrule{0em}{1pt}{1pt}
        \bottomrule
    \end{tabular}   
    \end{threeparttable}
\end{table}




\begin{figure}[t]
	\centering
	\includegraphics[width=8.8cm]{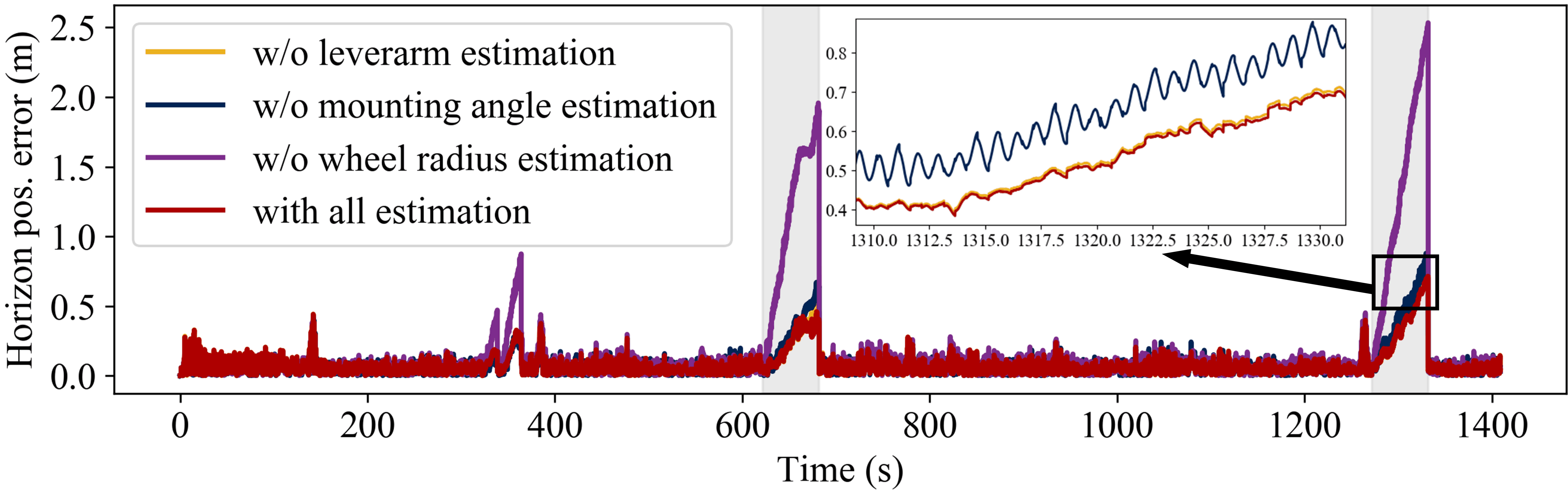}
	\caption{Horizontal position error drift of Wheel-GINS during \SI{60}{s} GNSS outages with and without online estimation of the Wheel-IMU installation parameters in Seq.~2.}
	\label{fig:all_ablation}
\end{figure}

\subsection{Online Wheel-IMU Installation Parameters Estimation}
In this section, we delve deeper into Wheel-GINS's capacity for online installation parameters estimation to give better insights into the system characteristics. The results support our second claim that Wheel-GINS can effectively estimate Wheel-IMU installation parameters, including the Wheel-IMU leverarm and mounting angle and the wheel radius scale error online, thus greatly improving the pose estimation accuracy.
\begin{figure}[t]
	\centering
	\includegraphics[width=8.8cm]{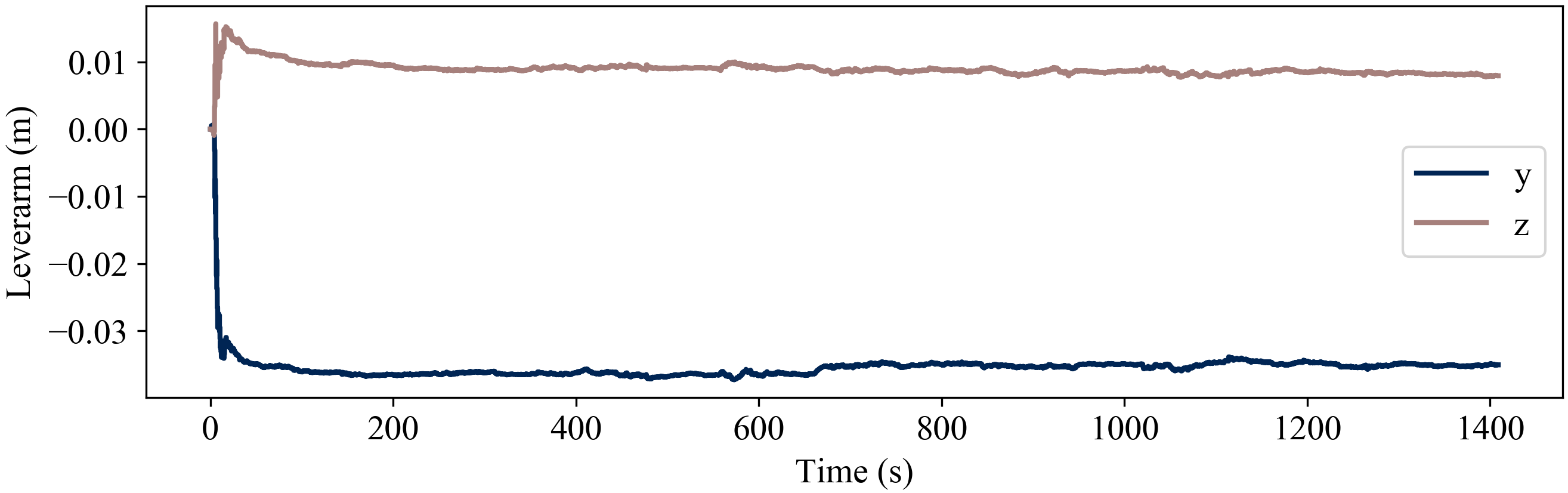}
	\caption{Online Wheel-IMU leverarm estimation results in Seq.~2. $y$ and $z$ represent the two leverarm components in the \textit{y}-axis and \textit{z}-axis of the \textit{body}-frame (see Eq.~2).}
	\label{fig:leverarm_estimation}
\end{figure}

First, we compare the positioning accuracy of Wheel-GINS during GNSS outages with and without the online estimation of the Wheel-IMU installation parameters to qualitatively illustrate the necessity of online estimation of the Wheel-IMU installation parameters. We set two \SI{60}{s} GNSS outages in Seq.~2. Fig.~\ref{fig:all_ablation} shows the horizon position error of Wheel-GINS with and without the online estimation of the Wheel-IMU installation parameters in Seq.~2. 

We can see that the positioning accuracy of Wheel-GINS is significantly improved with the online estimation of the Wheel-IMU installation parameters. Specifically, the influence of the wheel radius scale error is more evident in this experiment. This is because the wheel radius scale error directly affects the wheel velocity estimation, which is crucial when GNSS is unavailable. In addition, we can see that the Wheel-IMU leverarm and mounting angle errors also introduce positioning errors if not appropriately compensated. The Wheel-IMU mounting angle modulates a sine signal onto the positioning error because of the continuous rotation of the wheel. The influence of the Wheel-IMU leverarm is not significant in this experiment because it is less than \SI{5}{cm} (see Fig.~\ref{fig:leverarm_estimation}). We can constrain the Wheel-IMU leverarm error within this level by carefully installing the IMU. In conclusion, each Wheel-IMU installation error causes position error for Wheel-GINS at different levels if not calibrated properly. In the following section, we show experimental results to analyze the online estimation of each Wheel-IMU installation parameters in the proposed Wheel-GINS, respectively.

\subsubsection{Online Wheel-IMU Leverarm Estimation}
We first investigate the online estimation of the Wheel-IMU leverarm in Wheel-GINS. Because it is difficult to get the ground truth value of the Wheel-IMU leverarm, we can only evaluate the effectiveness of the leverarm online estimation by looking at the convergence of the error. Fig.~\ref{fig:leverarm_estimation} plots the Wheel-IMU leverarm estimation result in Seq.~2. The figure shows that the Wheel-IMU leverarm error can be effectively estimated in Wheel-GINS, which converges in around \SI{90}{s}. Because it is difficult to accurately measure the Wheel-IMU leverarm in practice, the online estimation of the Wheel-IMU leverarm is essential for Wheel-GINS to achieve high positioning accuracy.

\begin{figure}[t]
	\centering
	\includegraphics[width=8.8cm]{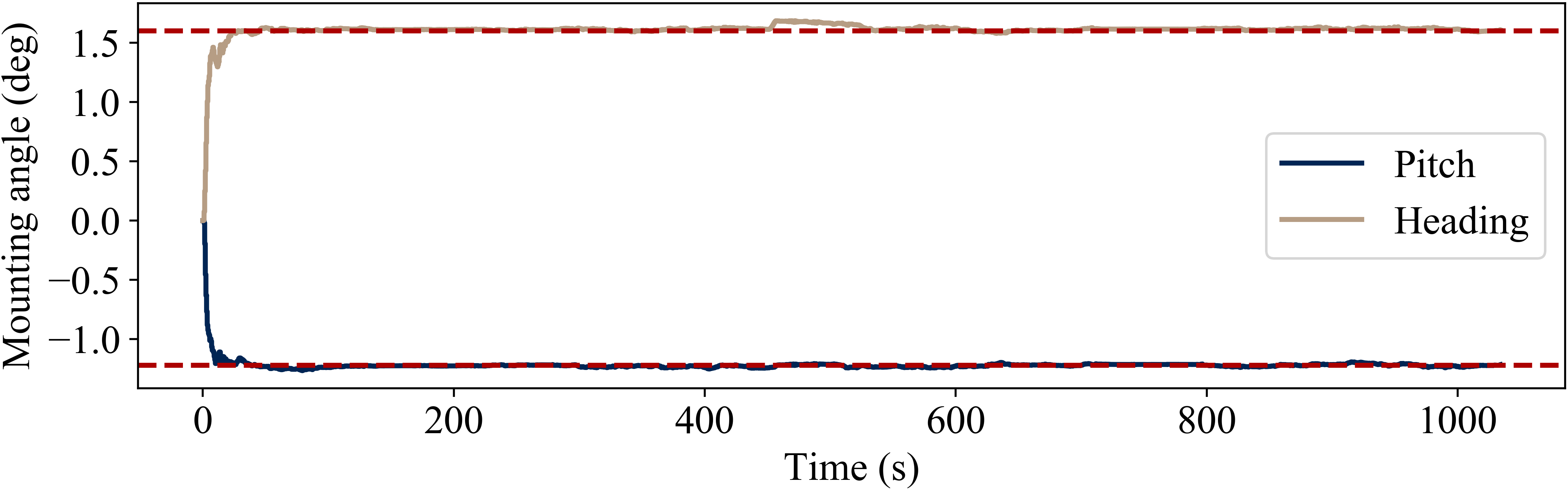}
    \caption{Online Wheel-IMU mounting angle estimation results in Seq.~3. The red dashed line represents the reference value calculated by the offline calibration method~\cite{chen2019} (pitch mounting angle: -\SI{1.22}{\degree}, heading mounting angle: \SI{1.60}{\degree}).}
	\label{fig:mounting_angle_err}
\end{figure}
\subsubsection{Online Wheel-IMU Mounting Angle Estimation}
Experimental results in Wheel-INS~\cite{niu2021} and Tan~\cite{tan2024sensorj} have illustrated that the Wheel-IMU mounting angle error causes significant pose error if it is not estimated and compensated properly. In this section, we analyze the online estimation of the Wheel-IMU mounting angle in Wheel-GINS. Fig.~\ref{fig:mounting_angle_err} compares the online estimation results of the Wheel-IMU mounting angle in Wheel-GINS with the offline calibration results~\cite{chen2019} in Seq.~3. We can see that the Wheel-IMU mounting angle error converges to the reference value in around \SI{30}{s} in Wheel-GINS. After convergence, it remains stable even when the vehicle occasionally turns. Fig.~\ref{fig:platforms_and_trajs} (f) shows that there are some turns and even u-turns in Seq.~3 because the vehicle traversed the same road back and forth twice. We can see from Fig.~\ref{fig:mounting_angle_err} that it does not disrupt the Wheel-IMU mounting angle estimation. These results back up our claim in {Section IV-B-3} that the Wheel-IMU mounting angle estimation converges fast in Wheel-GINS, and the occasional normal turning of the vehicle has negligible impact on the Wheel-IMU mounting angle estimation.

\begin{figure}[t]
	\centering
	\includegraphics[width=8.8cm]{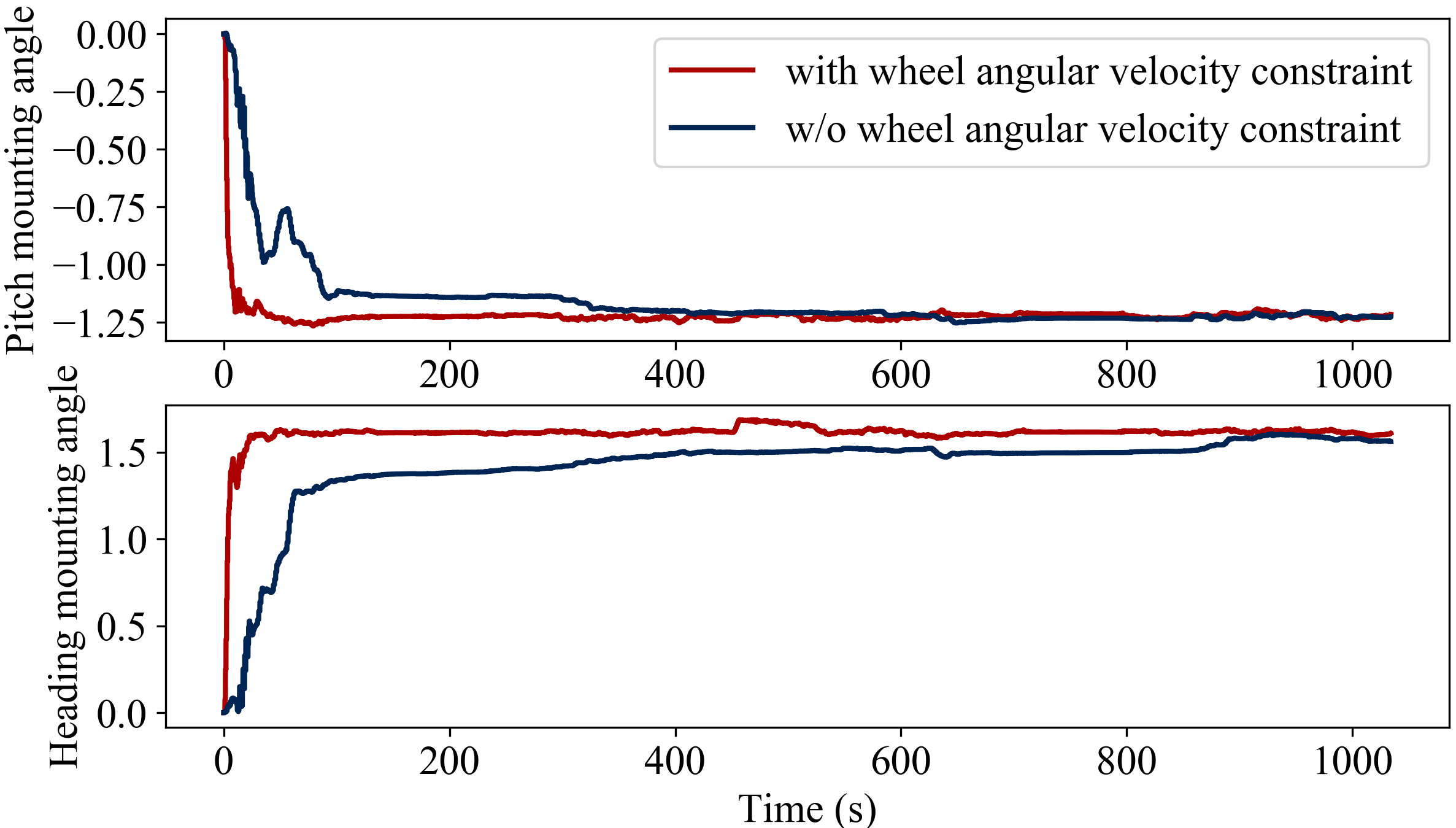}
	\caption{Online Wheel-IMU mounting angle estimation with/without the proposed wheel angular velocity constraint in Seq.~3. (Unit: degree).}
	\label{fig:angularvelocity_comparison}
\end{figure}

\begin{figure}[t]
	\centering
	\setlength{\abovecaptionskip}{0.cm}
	\includegraphics[width=8.8cm]{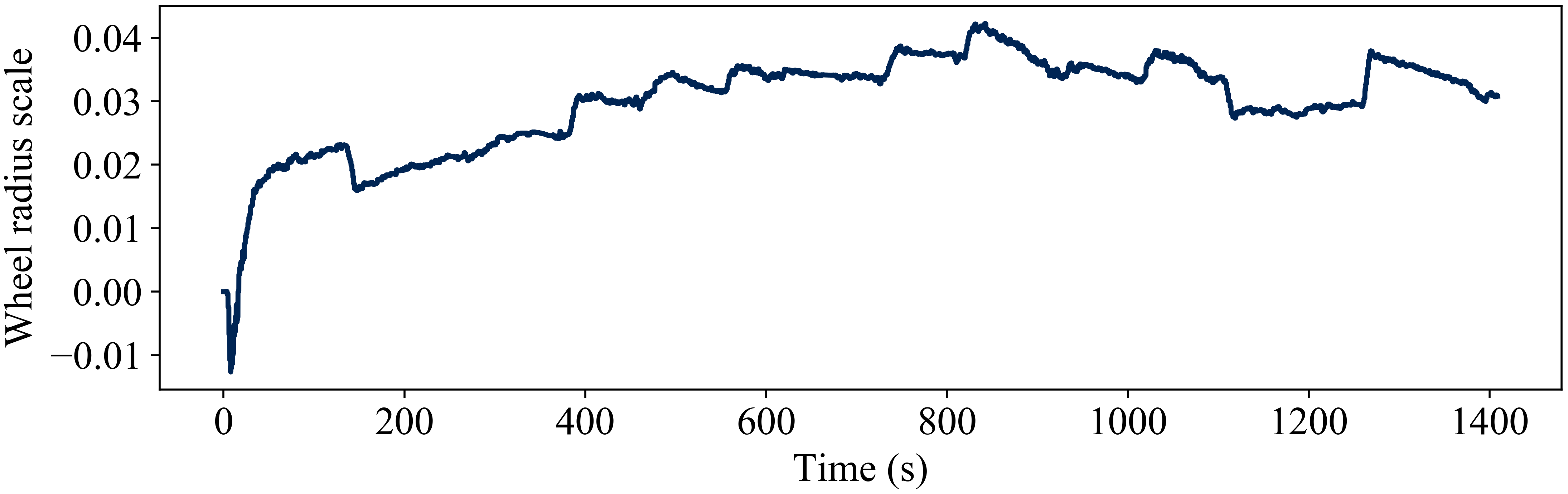}
	\caption{Online wheel radius scale error estimation results of Wheel-GINS in Seq.~2.}
	\label{fig:wheel_radius_err_std}
\end{figure}

One may argue that the GNSS position can also help estimate the Wheel-IMU heading misalignment as it provides absolute heading information for the vehicle; thus, it is unnecessary to integrate the proposed wheel angular velocity constraint model. We conduct a comparison experiment to show that the proposed angular velocity measurement can significantly accelerate the convergence of the Wheel-IMU mounting angle error. Fig.~\ref{fig:angularvelocity_comparison} shows the results. As we can see, with the integration of the proposed wheel angular velocity constraint, the convergence time of the Wheel-IMU pitch and heading mounting angle estimation has been reduced from around \SI{400}{s} and \SI{900}{s} to \SI{30}{s}, respectively.

Although the GNSS position indirectly reflects the vehicle heading, which helps to estimate the Wheel-IMU mounting angle, it takes a long time to achieve convergence. In addition, the accuracy of the GNSS position is also critical. When the accuracy of GNSS positioning is poor, such as in complex environments, it even impairs the estimation of the Wheel-IMU mounting angle. However, the proposed wheel angular velocity constraint is not affected by the environment. Therefore, the proposed wheel angular velocity measurement plays a key role in estimating the Wheel-IMU mounting angle to improve the pose accuracy in Wheel-GINS.

\subsubsection{Online Wheel Radius Scale Error Estimation}
We now conduct experiments to illustrate that the wheel radius scale error can be effectively estimated in Wheel-GINS. Fig.~\ref{fig:wheel_radius_err_std} plots the online estimation result of the wheel radius scale error in Seq.~2. From the figure, we can see that the wheel radius scale error drifts at the beginning because of the coupling effect from other installation parameters and IMU errors. Still, it soon converges to a reasonable range. Due to the pneumatic nature and softness of the tires in the trolley used in Seq.~2, deformation is prone to occur on uneven road surfaces. Consequently, there are some variations of the wheel radius scale estimation along with the sequence. It is also difficult to accurately calibrate the wheel radius scale error offline because it varies due to the terrain, tire pressure, vehicle weight, and so on. Therefore, the online estimation of the wheel radius scale error is necessary for Wheel-GINS to achieve high positioning accuracy, especially for the robots with pneumatic tires prone to deformation.

\subsection{Discussion}
In this paper, we propose to integrate GNSS with a Wheel-IMU to build a GNSS/INS integrated navigation system for long-term seamless state estimation of the wheeled robots. The proposed Wheel-GINS achieves similar positioning accuracy compared to the conventional ODO-GINS because they somewhat fuse the same information: egomotion estimation from the strapdown INS, vehicle velocity from either the odometer or Wheel-IMU, and absolute position from the GNSS. However, Wheel-GINS exhibits significant advantages over ODO-GINS during GNSS outages thanks to the inherent rotation modulation effect of the Wheel-IMU.

In addition to constraining the error drift of Wheel-INS, Wheel-GINS can effectively estimate the Wheel-IMU installation parameters, including the Wheel-IMU leverarm and mounting angle and the wheel radius scale error online. Experimental results have also illustrated the necessity and importance of the online estimation of Wheel-IMU installation parameters to improve position accuracy. Thanks to the reliable online estimation of the Wheel-IMU installation parameters, there is no need for prior calibration in Wheel-GINS, making Wheel-GINS a self-contained solution for wheeled robot state estimation and improving the system's applicability in practical applications.

\section{Conclusion}
Our goal in this study was to build a long-term accurate and robust localization system for wheeled robots. For that, we proposed Wheel-GINS, a GNSS/INS integrated navigation system using a wheel-mounted IMU. Based on Wheel-INS~\cite{niu2021}, we integrated the GNSS position observation into the EKF pipeline. To take full advantage of the absolute position information from GNSS, we augmented Wheel-IMU installation parameters, including the Wheel-IMU leverarm and mounting angle and the wheel radius scale error, into the state vector to be estimated online. Furthermore, we proposed a novel wheel angular velocity observation model to accelerate the convergence of the Wheel-IMU mounting angle error.

Real-world experimental results have illustrated that the proposed Wheel-GINS can achieve similar localization performance compared to the conventional ODO-GINS when GNSS is always available, while it significantly outperforms ODO-GINS during GNSS outages. Additionally, the Wheel-IMU installation parameters, including the Wheel-IMU leverarm and mounting angle and wheel radius scale error, can be effectively estimated online, thus improving the localization accuracy and the practicality of the system. 


\section{Acknowledgment}
We thank Liqiang Wang for his help in the comparison experiments with ODO-GINS and Markus Wagner for his help in conducting the experiments. 

\appendix
Here, we derive the matrix $\mathbf{A}$ in Eq.~\ref{vehicle_vel_obs_model}. Given the two Wheel-IMU mounting angles, $\theta_m$ (pitch mounting angle), and $\psi_m$ (heading mounting angle), the rotation matrix from the \textit{body}-frame to the \textit{wheel}-frame can be expressed as
\begin{equation}
\mathbf{R}_b^w = \begin{bmatrix}
\cos{\theta_{m}}\cos{\psi_{m}} & -\sin\psi_m & \sin{\theta_{m}}\cos{\psi_{m}}\\
\cos{\theta_{m}}\sin{\psi_{m}} & \cos\psi_m & \sin{\theta_{m}}\sin{\psi_{m}}\\
 -\sin{\theta_{m}} & 0  &  \cos{\theta_{m}}
\end{bmatrix}.
\end{equation}
In Eq.~\ref{vehicle_vel_obs_model}, ${\mathbf{R}}_{b(1,:)}^w$ indicates the first row of ${\mathbf{R}}_{b}^w$. We only use the first row because we only need the wheel angular velocity in the \textit{x}-axis to compute the wheel velocity. Because the estimated Wheel-IMU mounting angle contains errors, we have the estimated value of ${\mathbf{R}}_{b(1,:)}^w$ as
\begin{equation}
\begin{aligned}
	\hat{\mathbf{R}}_{b(1,:)}^w &= {\mathbf{R}}_{b(1,:)}^w + \delta{\mathbf{R}}_{b(1,:)}^w\\
	&= \begin{bmatrix}
	\cos{\hat{\theta}_{m}}\cos{\hat{\psi}_{m}} & -\sin\hat{\psi}_m & \sin{\hat{\theta}_{m}}\cos{\hat{\psi}_{m}}
	\end{bmatrix}\\
	&= \begin{bmatrix} 
		\cos({\theta}_{m}\!+\!\delta\theta_{m})\cos({\psi}_{m}\!+\!\delta\psi_{m}) \\ -\sin({\psi}_m\!+\!\delta\psi_{m}) \\ \sin({\theta}_{m}\!+\!\delta\theta_{m})\cos({\psi}_{m}\!+\!\delta\psi_{m})
	\end{bmatrix}^\top.
\end{aligned}
\end{equation}

After the expansion of the terms on the right side of the equation and ignoring the second-order small terms, we have
\begin{equation}
\begin{aligned}
	\delta{\mathbf{R}}_{b(1,:)}^w = \begin{bmatrix}
	-\sin\theta_m\cos\psi_m\delta\theta_m - \cos\theta_m\sin\psi_m\delta\psi_m\\
	-\cos\psi_m\delta\psi_m\\
	\cos\theta_m\cos\psi_m\delta\theta_m - \sin\theta_m\sin\psi_m\delta\psi_m
	\end{bmatrix}^\top.
\end{aligned}
\end{equation}

Then the coefficent matrix of the mounitng angle error $\mathbf{A}$ in Eq.~\ref{vehicle_vel_obs_model} can be expressed as
\begin{equation}
	{\mathbf{A}} = \begin{bmatrix}
		-\omega^b_x{r}\sin\theta_m\cos\psi_m\!+\!\omega^b_z{r}\cos\theta_m\cos\psi_m \\ 
		-\omega^b_x{r}\cos\theta_m\sin\psi_m\!-\!\omega^b_y{r}\cos\psi_m\!-\!\omega^b_z{r}\sin\theta_m\sin\psi_m 
		\end{bmatrix}^\top,
\end{equation}
where $\omega^b_x, \omega^b_y, \omega^b_z$ represent the components of the Wheel-IMU angular rate measurement $\bm{\omega}$ along the \textit{x}, \textit{y}, and \textit{z} axes, namely, $\bm{\omega} = \begin{bmatrix} \omega^b_x \!&\! \omega^b_y \!&\! \omega^b_z \end{bmatrix}^\top$.

\small
\bibliographystyle{IEEEtran}
\bibliography{bibliography/IEEEabrv, bibliography/my_abrv, bibliography/wheelgins}

\begin{thebibliography}{10}
\providecommand{\url}[1]{#1}
\csname url@samestyle\endcsname
\providecommand{\newblock}{\relax}
\providecommand{\bibinfo}[2]{#2}
\providecommand{\BIBentrySTDinterwordspacing}{\spaceskip=0pt\relax}
\providecommand{\BIBentryALTinterwordstretchfactor}{4}
\providecommand{\BIBentryALTinterwordspacing}{\spaceskip=\fontdimen2\font plus
\BIBentryALTinterwordstretchfactor\fontdimen3\font minus \fontdimen4\font\relax}
\providecommand{\BIBforeignlanguage}[2]{{%
\expandafter\ifx\csname l@#1\endcsname\relax
\typeout{** WARNING: IEEEtran.bst: No hyphenation pattern has been}%
\typeout{** loaded for the language `#1'. Using the pattern for}%
\typeout{** the default language instead.}%
\else
\language=\csname l@#1\endcsname
\fi
#2}}
\providecommand{\BIBdecl}{\relax}
\BIBdecl

\bibitem{wong2023tits}
C.~Wong, B.~Xia, Q.~Peng, W.~Yuan, and X.~You, ``{MSN}: Multi-style network for trajectory prediction,'' \emph{{IEEE} Transactions on Intelligent Transportation Systems}, vol.~24, no.~9, pp. 9751--9766, 2023.

\bibitem{xiaozhu2024tmc}
Z.~Xiao, J.~Shu, H.~Jiang, G.~Min, J.~Liang, and A.~Iyengar, ``Toward collaborative occlusion-free perception in connected autonomous vehicles,'' \emph{IEEE Transactions on Mobile Computing}, vol.~23, no.~5, pp. 4918--4929, 2024.

\bibitem{wong2024cvpr}
C.~Wong, B.~Xia, Z.~Zou, Y.~Wang, and X.~You, ``Socialcircle: Learning the angle-based social interaction representation for pedestrian trajectory prediction,'' in \emph{{Proc. of the IEEE Conf. on Computer Vision and Pattern Recognition (CVPR)}}, 2024, pp. 19\,005--19\,015.

\bibitem{pan2024tro}
Y.~Pan, X.~Zhong, L.~Wiesmann, T.~Posewsky, J.~Behley, and C.~Stachniss, ``{PIN-SLAM}: {LiDAR SLAM} using a point-based implicit neural representation for achieving global map consistency,'' \emph{{IEEE} Transactions on Robotics}, vol.~40, pp. 4045--4064, 2024.

\bibitem{lavalle2006planning}
S.~LaValle, \emph{Planning Algorithms}.\hskip 1em plus 0.5em minus 0.4em\relax Cambridge, United Kingdom: Cambridge University Press, 2006.

\bibitem{niu2021}
X.~Niu, Y.~Wu, and J.~Kuang, ``{Wheel-INS}: A wheel-mounted {MEMS} {IMU}-based dead reckoning system,'' \emph{{IEEE} Transactions on Vehicular Technology}, vol.~70, no.~10, pp. 9814--9825, 2021.

\bibitem{wu2021}
Y.~Wu, X.~Niu, and J.~Kuang, ``A comparison of three measurement models for the wheel-mounted {MEMS} {IMU}-based dead reckoning system,'' \emph{{IEEE} Transactions on Vehicular Technology}, vol.~70, no.~11, pp. 11\,193--11\,203, 2021.

\bibitem{savage2007strapdown}
P.~G. Savage, \emph{Strapdown Analytics, Part 1}.\hskip 1em plus 0.5em minus 0.4em\relax Minnesota, United States of America: Strapdown Associates, 2000.

\bibitem{Shin2005}
E.-H. Shin, ``Estimation techniques for low-cost inertial navigation,'' Ph.D. dissertation, Department of Geomatics Engineering, University of Calgary, Calgary, Canada, 2005.

\bibitem{wu2022ral}
Y.~Wu, J.~Kuang, X.~Niu, J.~Behley, L.~Klingbeil, and H.~Kuhlmann, ``{Wheel-SLAM}: Simultaneous localization and terrain mapping using one wheel-mounted {IMU},'' \emph{{IEEE} Robotics and Automation Letters}, vol.~8, no.~1, pp. 280--287, 2023.

\bibitem{li2022sana}
X.~Li, J.~Huang, X.~Li, Z.~Shen, J.~Han, L.~Li, and B.~Wang, ``Review of {PPP--RTK}: Achievements, challenges, and opportunities,'' \emph{Satellite Navigation}, vol.~3, no.~28, 2022.

\bibitem{naser2021sensors}
A.~A. Youssef, N.~Al-Subaie, N.~El-Sheimy, and M.~Elhabiby, ``Accelerometer-based wheel odometer for kinematics determination,'' \emph{Sensors}, vol.~21, no.~4, 2021.

\bibitem{gersdorf2013}
B.~Gersdorf and U.~Freese, ``A {Kalman} filter for odometry using a wheel mounted inertial sensor,'' in \emph{Proc. of the Int. Conf. on Informatics in Control, Automation and Robotics (ICINCO)}, 2013.

\bibitem{coulter2011}
E.~H. Coulter, P.~M. Dall, L.~Rochester, J.~P. Hasler, and M.~H. Granat, ``Development and validation of a physical activity monitor for use on a wheelchair,'' \emph{Spinal Cord}, vol.~49, no.~3, pp. 445--450, 2011.

\bibitem{collin2014tim}
J.~Collin, M.~Kirkko-Jaakkola, and J.~Takala, ``Effect of carouseling on angular rate sensor error processes,'' \emph{{IEEE} Transactions on Instrumentation and Measurement}, vol.~64, no.~1, pp. 230--240, 2014.

\bibitem{collin2014tvt}
J.~Collin, ``{MEMS IMU} carouseling for ground vehicles,'' \emph{{IEEE} Transactions on Vehicular Technology}, vol.~64, no.~6, pp. 2242--2251, 2014.

\bibitem{tan2024sensorj}
C.~Tan, ``Dead reckoning localization using a single wheel-mounted {IMU} with the simultaneous estimation of mounting parameters,'' \emph{{IEEE} Sensors Journal}, vol.~24, no.~3, pp. 3797--3810, 2024.

\bibitem{wu2009}
Y.~Wu, M.~Wu, X.~Hu, and D.~Hu, ``Self-calibration for land navigation using inertial sensors and odometer: Observability analysis,'' in \emph{Proc. of the AIAA Guidance, Navigation, and Control Conf.}, 2009.

\bibitem{wu2022tits}
Y.~Wu, J.~Kuang, and X.~Niu, ``{Wheel-INS2}: Multiple {MEMS IMU}-based dead reckoning system with different configurations for wheeled robots,'' \emph{{IEEE} Transactions on Intelligent Transportation Systems}, vol.~24, no.~3, pp. 3064--3077, 2023.

\bibitem{angrisano2010phd}
A.~Angrisano, ``{GNSS/INS} integration methods,'' Ph.D. dissertation, Department of Applied Sciences, Parthenope University of Naples, Naples, Italy, 2010.

\bibitem{wen2021navigation}
W.~Wen, T.~Pfeifer, X.~Bai, and L.-T. Hsu, ``Factor graph optimization for {GNSS/INS} integration: A comparison with the extended {Kalman} filter,'' \emph{NAVIGATION}, vol.~68, no.~2, pp. 315--331, 2021.

\bibitem{zhangquan2024tiv}
Q.~Zhang, H.~Lin, L.~Ding, Q.~Chen, T.~Zhang, and X.~Niu, ``{RANSAC}-based fault detection and exclusion algorithm for single-difference tightly coupled {GNSS/INS} integration,'' \emph{{IEEE} Transactions on Intelligent Vehicles}, vol.~9, no.~2, pp. 3986--3997, 2024.

\bibitem{xiaozhu2020tits}
V.~Havyarimana, Z.~Xiao, A.~Sibomana, D.~Wu, and J.~Bai, ``A fusion framework based on sparse gaussian–wigner prediction for vehicle localization using {GDOP} of {GPS} satellites,'' \emph{IEEE Transactions on Intelligent Transportation Systems}, vol.~21, no.~2, pp. 680--689, 2020.

\bibitem{zhang2022ral}
H.~Zhang, X.~Xia, M.~Nitsch, and D.~Abel, ``Continuous-time factor graph optimization for trajectory smoothness of {GNSS/INS} navigation in temporarily {GNSS}-denied environments,'' \emph{{IEEE} Robotics and Automation Letters}, vol.~7, no.~4, pp. 9115--9122, 2022.

\bibitem{hua2023sana}
T.~Hua, L.~Pei, T.~Li, J.~Yin, G.~Liu, and W.~Yu, ``{M2C-GVIO}: motion manifold constraint aided {GNSS}-visual-inertial odometry for ground vehicles,'' \emph{Satellite Navigation}, vol.~4, no.~13, 2023.

\bibitem{wen2019tvt}
W.~Wen, X.~Bai, Y.~C. Kan, and L.-T. Hsu, ``Tightly coupled {GNSS/INS} integration via factor graph and aided by fish-eye camera,'' \emph{{IEEE} Transactions on Vehicular Technology}, vol.~68, no.~11, pp. 10\,651--10\,662, 2019.

\bibitem{wang2020tim}
D.~Wang, Y.~Dong, Z.~Li, Q.~Li, and J.~Wu, ``Constrained {MEMS}-based {GNSS/INS} tightly coupled system with robust {Kalman} filter for accurate land vehicular navigation,'' \emph{{IEEE} Transactions on Instrumentation and Measurement}, vol.~69, no.~7, pp. 5138--5148, 2020.

\bibitem{xiaozhu2022tits}
Z.~Xiao, Y.~Chen, M.~Alazab, and H.~Chen, ``Trajectory data acquisition via private car positioning based on tightly-coupled {GPS/OBD} integration in urban environments,'' \emph{IEEE Transactions on Intelligent Transportation Systems}, vol.~23, no.~7, pp. 9680--9691, 2022.

\bibitem{loeliger2004factorgraph}
H.-A. Loeliger, ``An introduction to factor graphs,'' \emph{{IEEE} Signal Processing Magazine}, vol.~21, no.~1, pp. 28--41, 2004.

\bibitem{suzuki2024ral}
T.~Suzuki, ``Attitude-estimation-free {GNSS} and {IMU} integration,'' \emph{{IEEE} Robotics and Automation Letters}, vol.~9, no.~2, pp. 1090--1097, 2024.

\bibitem{zhang2024tiv}
L.~Zhang, W.~Wen, L.-T. Hsu, and T.~Zhang, ``An improved inertial preintegration model in factor graph optimization for high accuracy positioning of intelligent vehicles,'' \emph{{IEEE} Transactions on Intelligent Vehicles}, vol.~9, no.~1, pp. 1641--1653, 2024.

\bibitem{niu2023ral}
X.~Niu, H.~Tang, T.~Zhang, J.~Fan, and J.~Liu, ``{IC-GVINS}: A robust, real-time, {INS}-centric {GNSS}-visual-inertial navigation system,'' \emph{{IEEE} Robotics and Automation Letters}, vol.~8, no.~1, pp. 216--223, 2023.

\bibitem{chi2023ral}
C.~Chi, X.~Zhang, J.~Liu, Y.~Sun, Z.~Zhang, and X.~Zhan, ``{GICI-LIB}: A {GNSS}/{INS}/camera integrated navigation library,'' \emph{{IEEE} Robotics and Automation Letters}, vol.~8, no.~12, pp. 7970--7977, 2023.

\bibitem{li2022ral}
X.~Li, S.~Li, Y.~Zhou, Z.~Shen, X.~Wang, X.~Li, and W.~Wen, ``Continuous and precise positioning in urban environments by tightly coupled integration of {GNSS}, {INS} and vision,'' \emph{{IEEE} Robotics and Automation Letters}, vol.~7, no.~4, pp. 11\,458--11\,465, 2022.

\bibitem{liu2024tiv}
X.~Liu, W.~Wen, and L.-T. Hsu, ``{GLIO}: Tightly-coupled {GNSS}/{LiDAR}/{IMU} integration for continuous and drift-free state estimation of intelligent vehicles in urban areas,'' \emph{{IEEE} Transactions on Intelligent Vehicles}, vol.~9, no.~1, pp. 1412--1422, 2024.

\bibitem{li2022iot}
S.~Li, S.~Wang, Y.~Zhou, Z.~Shen, and X.~Li, ``Tightly coupled integration of {GNSS}, {INS}, and {LiDAR} for vehicle navigation in urban environments,'' \emph{{IEEE} Internet of Things Journal}, vol.~9, no.~24, pp. 24\,721--24\,735, 2022.

\bibitem{chang2024tiv}
J.~Chang, Y.~Zhang, S.~Fan, F.~Huang, D.~Xu, and L.-T. Hsu, ``An anti-spoofing model based on {MVM} and {MCCM} for a loosely-coupled {GNSS}/{INS}/{LiDAR} {Kalman} filter,'' \emph{{IEEE} Transactions on Intelligent Vehicles}, vol.~9, no.~1, pp. 1744--1755, 2024.

\bibitem{niu2007nav}
X.~Niu, S.~Nassar, and N.~EI-Sheimy, ``An accurate land-vehicle {MEMS IMU/GPS} navigation system using 3d auxiliary velocity updates,'' \emph{NAVIGATION}, vol.~54, no.~3, pp. 177--188, 2007.

\bibitem{Zhang2021mst}
Z.~Zhang, X.~Niu, H.~Tang, Q.~Chen, and T.~Zhang, ``{GNSS/INS/ODO}/wheel angle integrated navigation algorithm for an all-wheel steering robot,'' \emph{Measurement Science and Technology}, vol.~32, no.~11, pp. 115--122, 2021.

\bibitem{ouyang2020}
W.~Ouyang, Y.~Wu, and H.~Chen, ``{INS}/odometer land navigation by accurate measurement modeling and multiple-model adaptive estimation,'' \emph{{IEEE} Transactions on Aerospace and Electronic Systems}, vol.~57, no.~1, pp. 245--262, 2021.

\bibitem{dissanayake2001}
G.~Dissanayake, S.~Sukkarieh, E.~Nebot, and H.~Durrant-Whyte, ``The aiding of a low-cost strapdown inertial measurement unit using vehicle model constraints for land vehicle applications,'' \emph{{IEEE} Transactions on Robotics and Automation}, vol.~17, no.~5, pp. 731--747, 2001.

\bibitem{zhangquan2020}
Q.~Zhang, Y.~Hu, and X.~Niu, ``Required lever arm accuracy of non-holonomic constraint for land vehicle navigation,'' \emph{{IEEE} Transactions on Vehicular Technology}, vol.~69, no.~8, pp. 8305--8316, 2020.

\bibitem{martin2020tiv}
M.~Brossard, A.~Barrau, and S.~Bonnabel, ``{AI-IMU} dead-reckoning,'' \emph{{IEEE} Transactions on Intelligent Vehicles}, vol.~5, no.~4, pp. 585--595, 2020.

\bibitem{Groves2013}
P.~D. Groves, \emph{Principles of {GNSS}, Inertial, and Multisensor Integrated Navigation Systems}.\hskip 1em plus 0.5em minus 0.4em\relax London, United Kingdom: Artech House, 2013.

\bibitem{chen2023ieeesj}
Q.~Chen, H.~Lin, J.~Kuang, Y.~Luo, and X.~Niu, ``Rapid initial heading alignment for {MEMS} land vehicular {GNSS/INS} navigation system,'' \emph{{IEEE} Sensors Journal}, vol.~23, no.~7, pp. 7656--7666, 2023.

\bibitem{chen2021}
Q.~Chen, Q.~Zhang, and X.~Niu, ``Estimate the pitch and heading mounting angles of the {IMU} for land vehicular {GNSS/INS} integrated system,'' \emph{{IEEE} Transactions on Intelligent Transportation Systems}, vol.~22, no.~10, pp. 6503--6515, 2021.

\bibitem{niu2023tmech}
X.~Niu, Y.~Peng, Y.~Dai, Q.~Chen, C.~Guo, and Q.~Zhang, ``Camera-based lane-aided multi-information integration for land vehicle navigation,'' \emph{{{IEEE/ASME} Transactions on Mechatronics}}, vol.~28, no.~1, pp. 152--163, 2023.

\bibitem{chang2019rs}
L.~Chang, X.~Niu, T.~Liu, J.~Tang, and C.~Qian, ``{GNSS/INS/LiDAR-SLAM} integrated navigation system based on graph optimization,'' \emph{Remote Sensing}, vol.~11, no.~9, 2019.

\bibitem{zhang2024sj}
T.~Zhang, M.~Yuan, L.~Wang, H.~Tang, and X.~Niu, ``A robust and efficient {IMU} array/{GNSS} data fusion algorithm,'' \emph{IEEE Sensors Journal}, vol.~24, no.~16, pp. 26\,278--26\,289, 2024.

\bibitem{xiaozhu2018if}
V.~Havyarimana, D.~Hanyurwimfura, P.~Nsengiyumva, and Z.~Xiao, ``A novel hybrid approach based-{SRG} model for vehicle position prediction in multi-{GPS} outage conditions,'' \emph{Information Fusion}, vol.~41, pp. 1--8, 2018.

\bibitem{chen2019}
Q.~Chen, X.~Niu, J.~Kuang, and J.~Liu, ``{IMU} mounting angle calibration for pipeline surveying apparatus,'' \emph{{IEEE} Transactions on Instrumentation and Measurement}, vol.~69, no.~4, pp. 1765--1774, 2019.

\end{thebibliography}

\end{document}